\documentclass{article}

\usepackage{PRIMEarxiv}

\errorcontextlines\maxdimen

\usepackage[utf8]{inputenc} % allow utf-8 input
\usepackage[T1]{fontenc}    % use 8-bit T1 fonts
\usepackage{url}            % simple URL typesetting
\usepackage{booktabs}       % professional-quality tables
\usepackage{nicefrac}       % compact symbols for 1/2, etc.
\usepackage{microtype}      % microtypography
\usepackage{fancyhdr}       % header
\usepackage{graphicx}
\usepackage{xcolor}

\usepackage{lineno,hyperref}
\hypersetup{colorlinks, citecolor=blue, linkcolor=red, urlcolor=green}
\usepackage{amsmath,amsfonts,amssymb}
\usepackage{graphics}
\usepackage{bm}
\usepackage{multirow}
\usepackage{arydshln}
\usepackage{caption}
\usepackage{subcaption}
\usepackage{threeparttable}
\usepackage{algorithm}
\usepackage{algpseudocode}

\title{Discovering interpretable Lagrangian of dynamical systems from data}

\author{Tapas Tripura\\
  Department of Applied Mechanics\\
  Indian Institute of Technology Delhi\\
  \texttt{tapas.t@am.iitd.ac.in} \\
  \And
      Souvik Chakraborty \\
  Department of Applied Mechanics\\
  Yardi School of Artificial Intelligence (ScAI)\\
  Indian Institute of Technology Delhi\\
  \texttt{souvik@am.iitd.ac.in} \\
}

\begin{document}
\maketitle

\begin{abstract}
A complete understanding of physical systems requires models that are accurate and obeys natural conservation laws. Recent trends in representation learning involve learning Lagrangian from data rather than the direct discovery of governing equations of motion. The generalization of equation discovery techniques has huge potential; however, existing Lagrangian discovery frameworks are black-box in nature. This raises a concern about the reusability of the discovered Lagrangian. In this article, we propose a novel data-driven machine-learning algorithm to automate the discovery of interpretable Lagrangian from data. The Lagrangian are derived in interpretable forms, which also allows the automated discovery of conservation laws and governing equations of motion. 
The architecture of the proposed framework is designed in such a way that it allows learning the Lagrangian from a subset of the underlying domain and then generalizing for an infinite-dimensional system. The fidelity of the proposed framework is exemplified using examples described by systems of ordinary differential equations and partial differential equations where the Lagrangian and conserved quantities are known.
\end{abstract}

% keywords can be removed

\section{Introduction}
Physical and engineering systems are governed by certain conservation laws, which are deeply rooted in the underlying symmetries of natural laws. The theories laid by Emmy Noether provide us with a profound understanding of the connections between conservation laws and underlying symmetries \cite{sardanashvily2016noether}. In works by Noether, the role of Lagrangian formalism (systems described by a Lagrangian) is overwhelming since most of the proof of the theorem is taken care of by the Lagrangian. The Lagrangian of a physical system encodes succinctly all the important information
needed for discovering the underlying physics \cite{calkin1996lagrangian} and conservation law and has application in different disciplines ranging from quantum field theory to electromagnetism, to continuum mechanics, to the general theory of relativity \cite{landau2013course,landau2013classical}. In this paper, we propose an algorithm that can discover interpretable Lagrangian from data and use it for discovering governing physics and conservation laws. 

Learning the Lagrangian of systems from data has gained some popularity in recent times. Due to the significant developments in data-driven \cite{rudy2017data,wu2020data,TRIPURA2023115783}, and physics-informed \cite{raissi2019physics,goswami2020transfer,tartakovsky2020physics,chakraborty2021transfer} neural network algorithms, researchers have suggested using neural network to extract Lagrangian from data. Initial works related to the discovery of Lagrangian can be linked to Hamiltonian Neural Networks (HNN) \cite{greydanus2019hamiltonian,toth2019hamiltonian}. Learning Hamiltonian from data using graph neural network can be found in \cite{sanchez2019hamiltonian}. In these approaches, the Hamiltonian is parameterized through the neural network, and a physics-informed neural network is constituted to learn the Hamiltonian from data. The limitation behind these networks was that Hamiltonian formalism requires that the coordinates of the system must be canonical, which is often restrictive for real-life systems. As an alternative, the Deep Lagrangian Networks (DeLaN) was proposed in \cite{lutter2019deep}. For learning the Hamiltonian and Lagrangian directly in Cartesian coordinates, Constrained Hamiltonian Neural Networks (CHNNs) and Constrained Lagrangian Neural Networks (CLNNs) were proposed in \cite{gruver2022deconstructing}. Motivated by the recent developments in graph neural networks, the Lagrangian graph neural network (LGNN) for learning the Lagrangian of rigid body dynamics was proposed in \cite{bhattoolearning}. However, these works were mostly limited to rigid body dynamics. Another elegant approach for discovering the Lagrangian of non-rigid body motions was proposed in \cite{cranmer2020lagrangian}. For a brief comparison of the similarities, differences, and the theory behind all the above frameworks, readers can further refer to \cite{zhong2021benchmarking}. The application of Hamiltonian and Lagrangian formalism for robotic control can also be found in \cite{gupta2019general,duong2021hamiltonian}.

One issue that arises in the above neural network-based frameworks is that the discovered Lagrangian models are not interpretable, and this significantly reduces its reusability. For instance, it is not possible to discover the governing equation from the discovered Lagrangian. Another issue with such frameworks resides in the fact that to learn the Lagrangian accurately multiple-time history of system responses is required, which is prohibitive from a practical point of view. From here, we deviate ourselves and focus on the recently published Sparse Identification of Nonlinear Dynamics (SINDy) algorithm \cite{brunton2016discovering}. The SINDy is a sparse system identification algorithm that automates the discovery of governing equations of motion from data. The advantage of the SINDy algorithm over its neural network counterparts \cite{rudy2019deep,raissi2018deep} is that it provides an interpretable form of the exact governing dynamics. The exact physics helps in the generalization of the discovered model to unseen environmental conditions. To discover the exact interpretable physics, it employs a sequential threshold least-square regression. Further, SINDy is computationally efficient and scalable with an increase in the dimension of the measurement states. The Bayesian approach for discovering governing physics from data can also be found in \cite{nayek2021spike,tripura2022model,tripura2023sparse}. These approaches, although robust to noise, are computationally demanding as compared to SINDy.

We hereby propose a data-driven framework for systematic, automatic, and accelerated discovery of Lagrangian from data. In particular, we leverage the well-established Lagrangian formalism to discover exact physics and the sequential threshold least-squares to discover the exact sparse Lagrangian of systems from data. The proposed framework is more informative than identifying the ordinary and partial differential equations (ODE/PDE) from data \cite{rudy2017data,rudy2019deep}. In comparison to ODE/PDE discovery, the proposed framework can also be automated using Legendre transformation through symbolic derivatives to automatically discover the conservation laws. In terms of data requirement, the proposed framework needs only a single observation of the system responses to accurately distills the Lagrangian of the underlying physical system.
The proposed approach has several key features that can be encapsulated into the following four points:
\begin{itemize}
    \item \textbf{Automated discovery of an exact interpretable form of Lagrangian}: The proposed approach discovers the exact interpretable form of Lagrangian of systems from data alone. Since the discovered Lagrangian is exact, therefore upon application of Noether's theorem and Legendre transformation, it provides the underlying conservation laws. From the discovery of Lagrangian to conservation laws, the proposed framework utilizes only a single observation of system responses.
    \item \textbf{Automated discovery of governing equations}: The proposed framework constrains the discovery of Lagrangian using the principle of minimal action. As a result, the discovered Lagrangian automatically satisfies the Euler-Lagrangian equations. Thus, the governing equations of motion of underlying systems are simultaneously discovered without human intervention.
    \item \textbf{Generalization to high-dimensional systems}: In the case of interaction between degrees-of-freedom in a high-dimensional system, the proposed approach learns the Lagrangian from the interaction of a relatively small subset of the system and can generalize the learned physics to the complete domain.
    \item \textbf{Zero-shot generalization and perpetual predictive ability}: The discovered Lagrangian satisfy the exact physics of the underlying systems. Therefore, the discovered equations of motion share the same predictive ability as the actual system. Thus in unseen environmental conditions, the discovered physics can be used to predict an infinite period of duration without affecting accuracy. 
\end{itemize}

The remainder of the paper is arranged as follows: in section \ref{sec:problem}, the problem statement is given. In section \ref{sec:methods}, the proposed data-driven framework for the discovery of the Lagrangian from data is briefly presented. In section \ref{sec:numerical}, numerical experiments are undertaken to showcase the novelty of the proposed data-driven framework. In section \ref{sec:conclusion}, the salient features of the proposed framework are revisited, and finally, the paper is concluded.

\section{Problem formulation}\label{sec:problem}
In this section, we formally define the problem statement of our proposed work.
Consider an $m$ degrees of freedom dynamical system with generalized coordinates $\{X_i; i=1, \ldots, m \}$, where the generalized coordinates refer to any set of independent coordinates. Further, we denote the kinetic and potential energy of this system by $T \in \mathbb{R}$ and $U \in \mathbb{R}$, respectively. The potential energy $U$ typically depends on the position of the system described by the generalized coordinates $\{X_i; i=1, \ldots, m \}$, whereas the kinetic energy $T$ can depend on both generalized displacement and velocity $\{\dot{X}_i; i=1, \ldots, m \}$. With this preliminary description, the scalar notion of kinetic energy $T$ and potential energy $U$ is expressed as $T \triangleq T(X_1, \ldots, X_m, \dot{X}_1, \ldots, \dot{X}_m)$, and $U \triangleq U(X_1, \ldots, X_m)$.
If a system is conservative, the differential $d(T+U)$ of the sum of kinetic energy $T$ and potential energy $U$ will be equal to zero, which is basically a statement of the principle of conservation of energy \cite{calkin1996lagrangian,brizard2014introduction}. When differentials of the sum of the kinetic energy $T$ and the potential energy $U$ over all the generalized coordinates are taken, we get the following representation of $T$ and $U$,
\begin{subequations}
    \begin{equation}\label{eq:dT}
        dT \triangleq \sum_{i=1}^m \frac{\partial}{\partial {X}_i} T\left(X_1, \ldots, X_m, \dot{X}_1, \ldots, \dot{X}_m\right) d {X}_i + \sum_{i=1}^m \frac{\partial}{\partial \dot{X}_i} T\left(X_1, \ldots, X_m, \dot{X}_1, \ldots, \dot{X}_m\right) d \dot{X}_i 
    \end{equation}
    \begin{equation}\label{eq:dU}
        dU \triangleq \sum_{i=1}^m \frac{\partial}{\partial X_i} U\left(X_1, X_2, \ldots, X_m\right) d X_i .
    \end{equation}
\end{subequations}
In the above equations, the differentials of the kinetic energy $T$ and the potential energy $U$ depend on the perturbations $d \dot{X}_i$ and $d X_i$. The dependence on the perturbations $d X_i$ can be relaxed by writing the equation of kinetic energy in generalized coordinates.
\begin{equation}
    T= \frac{1}{2} \sum_{i=1}^{n} \sum_{j=1}^{n} m_{ij} \dot{X}_i\dot{X}_j = \frac{1}{2} \sum_{i=1}^n \frac{\partial T}{\partial \dot{X}_i} \dot{X}_i,
\end{equation}
where $m_{ij} = \sum _k^n m_k (\partial r_k / \partial X_j) \cdot (\partial r_k / \partial X_j)$ is the coefficients of generalized mass matrix and $r_k \in \mathbb{R}^{n}$ is the position vector of each degree of freedom. Note that at the material level, the $n$-degrees of freedom system can be represented using an $n$-particle system. Applying the product rule in the above equation, one can get,
\begin{equation}\label{eq:product}
    dT= \frac{1}{2}\sum_{i=1}^m d\left(\frac{\partial T}{\partial \dot{X}_i}\right) \dot{X}_i + \frac{1}{2}\sum_{i=1}^m \frac{\partial T}{\partial \dot{X}_i} d \dot{X}_i .
\end{equation}
Subtracting the above equation from Eq. \eqref{eq:dT} yields,
\begin{equation}\label{eq:dT_final}
    d T=\sum_{i=1}^m \frac{d}{d t}\left(\frac{\partial T}{\partial \dot{X}_i}\right) d X_i - \sum_{i=1}^m \frac{\partial T}{\partial {X}_i} d {X}_i .
\end{equation}
From Eq. \eqref{eq:dT_final} and \eqref{eq:dU}, the differential $d(T+U)$ of the kinetic energy and potential energy can be obtained as follows,
\begin{equation}
    d(T+U)=\sum_{i=1}^m\left[\frac{d}{d t}\left(\frac{\partial T}{\partial \dot{X}_i}\right) - \frac{\partial T}{\partial {X}_i} + \frac{\partial U}{\partial X_i}\right] d X_i .
\end{equation}
Since the differential $d(T+U)$ must be zero, the only way it is possible is if the sum inside the square bracket is zero, i.e.,
\begin{equation}
    \frac{d}{d t}\left(\frac{\partial T}{\partial \dot{X}_i}\right) - \frac{\partial T}{\partial {X}_i} + \frac{\partial U}{\partial X_i}=0; \quad i=1,2, \ldots, m .
\end{equation}
We can further note that ${\partial U}/{\partial \dot{X}_i} =0$. Using these results, we can straightforwardly obtain Lagrange's equation as follows,
\begin{equation}\label{eq:lagrange_hom}
    \frac{d}{d t}\left(\frac{\partial \mathcal{L}}{\partial \dot{X}_i}\right)-\frac{\partial \mathcal{L}}{\partial X_i} = 0, \quad i=1,2, \ldots, m ,
\end{equation}
where $\mathcal{L}({{X}_i}, \dot{X}_i ) \in \mathbb{R} \triangleq T({X}_i, \dot{X}_i) -U({X}_i)$ is the Lagrangian. Equation \eqref{eq:lagrange_hom} represents Lagrange's equation for a conservative system, i.e., systems with constant energy. For nonconservative systems, Lagrange's equation is generalized by including a constraining potential $V_i$ as follows,
\begin{equation}\label{eq:lagrange_nonhom}
    \frac{d}{d t}\left(\frac{\partial \mathcal{L}}{\partial \dot{X}_i}\right)-\frac{\partial \mathcal{L}}{\partial X_i}=V_i, \quad i=1,2, \ldots, m .
\end{equation}
With this setup, given that we can obtain the measurements for system states $\mathbf{X} = [ {\bm{X}}, \dot{\bm{X}} ]$, we aim (i) to automate the discovery of the exact analytical form of the Lagrangian $\mathcal{L}({\bm{X}}, \dot{\bm{X}} )$ by constraining the Lagrangian to satisfy Eqs. \eqref{eq:lagrange_hom} and \eqref{eq:lagrange_nonhom}, (ii) to automate the discovery of the governing equations of motion using the discovered Lagrangian $\mathcal{L}({\bm{X}}, \dot{\bm{X}} )$, which will have perpetual prediction capability, and (iii) to automate the discovery of conservation laws using the principle of energy conservation on discovered Lagrangian $\mathcal{L}({\bm{X}}, \dot{\bm{X}} )$. To achieve tasks (i)--(iii), we restrict our access to a single trajectory of the system states ${\bm{X}} \in \mathbb{R}^{m}$ and $\dot{\bm{X}} \in \mathbb{R}^{m}$ only.

\section{Learning of Lagrangian of physical systems from state measurement}\label{sec:methods}
The expression of the Lagrangian defined in the previous section entirely depends on the form of internal and external energy potentials. The typical forms of these potentials are sparse in 
nature, i.e., the resulting analytical expression of the Lagrangian will contain only a few most relevant physical terms. We embed this information in the proposed framework and leverage the concept of sparse regression \cite{brunton2016discovering,boninsegna2018sparse} to automate the discovery of the Lagrangian from field observation data.
Towards discovering the Lagrangian from data, we first gather the time history of displacement and velocity states for $t \in [0,T]$ at some discrete time step $\Delta t$, where $T$ is the total duration of the measurement period. Note that the states can also be angular displacement and angular velocities. 
Assuming that the filtration of states $\mathcal{F}^{X}$ contains $N$-time steps, and there are $m$-system states, we construct a matrix structure of the measurements as follows,
\begin{equation}\label{eq:measurement}
    \mathbf{X} = \left[ {\begin{array}{*{20}{c}}
    {X_{1_1}} & {X_{2_1}} & \cdots & {X_{m_1}}\\
    {X_{1_2}} & {X_{2_2}} & \cdots & {X_{m_2}}\\
     \vdots & \vdots & \ddots & \vdots \\
    {X_{1_N}} & {X_{2_N}} & \cdots & {X_{m_N}}\\
    \end{array}} \right]; \quad 
    \dot{\mathbf{X}} = \left[ {\begin{array}{*{20}{c}}
    {\dot{X}_{1_1}} & {\dot{X}_{2_1}} & \cdots & {\dot{X}_{m_1}}\\
    {\dot{X}_{1_2}} & {\dot{X}_{2_2}} & \cdots & {\dot{X}_{m_2}}\\
     \vdots & \vdots & \ddots & \vdots \\
    {\dot{X}_{1_N}} & {\dot{X}_{2_N}} & \cdots & {\dot{X}_{m_N}}\\
    \end{array}} \right].
\end{equation}

Next, we assume that the Lagrangian $\mathcal{L}({\mathbf{X}}, \dot{\mathbf{X}} )$ can be expressed as a weighted linear superposition of certain candidate basis functions \cite{brunton2016discovering,nayek2021spike}. Since the choice of candidate functions can be arbitrary, we construct a symbolic dictionary $\mathbf{D}(\mathbf{X}, \dot{\mathbf{X}})$ of all the possible choices of energy potentials. The possible choices for candidate potential functions can be polynomials of states, harmonics of states, differences of states, polynomials of differences of states, harmonics of differences of states, etc. The functions of $\mathbf{D}(\mathbf{X}, \dot{\mathbf{X}})$ are mapping of each column of $\mathbf{X}$ and $\dot{\mathbf{X}}$. A demonstration of the dictionary employed in this work is given below, 
\begin{equation}\label{eq:library}
    \mathbf{D}(\mathbf{X}, \dot{\mathbf{X}}) = \left[ {\begin{array}{*{20}{c}}
            {\bf{1}} & {{{\rm P}^{\rho}} \left({\mathbf{X}}, \dot{\mathbf{X}}\right)} & \cdots & {\mathop{\rm sin}\left({\mathbf{X}}, \dot{\mathbf{X}}\right)} & \cdots & {{{\rm P}^{\rho}}\left(\left|{\mathbf{X}}- \dot{\mathbf{X}} \right| \right)} & \cdots & {\mathop{\rm sin}\left(\left|{\mathbf{X}}- \dot{\mathbf{X}} \right| \right)} & \cdots 
    \end{array}} \right],
\end{equation}
where $\rho$ denotes the degree of polynomials. We then evaluate the above symbolic dictionary on the state measurements in  \eqref{eq:measurement}, where each column of the $\mathbf{D}(\mathbf{X}, \dot{\mathbf{X}})$ represents a possible candidate to be included in the final model of the $\mathcal{L}({\mathbf{X}}, \dot{\mathbf{X}} )$. For upcoming discussions, we assume that there are a total $K$-basis functions and represent the dictionary functions by $\ell_k; k=1,\ldots, K$. Using the above description, we express the Lagrangian ($\mathcal{L}$) for $i^{th}$-DOF as,
\begin{equation}\label{eq:lagrange_lib}
    \mathcal{L}({\bm{X}_i}, \dot{\bm{X}}_i ) = \theta_1 \ell_1({\bm{X}_i}, \dot{\bm{X}}_i ) + \theta_2 \ell_2({\bm{X}_i}, \dot{\bm{X}}_i ) + \ldots + \theta_K \ell_K({\bm{X}_i}, \dot{\bm{X}}_i ) ,
    % \mathcal{L}({X_i}, \dot{X}_i ) = \theta_{i_1} \ell_1({X_i}, \dot{X}_i) + \theta_{i_2} \ell_2({X_i}, \dot{X}_i) + \cdots + \theta_{i_K} \ell_K({X_i}, \dot{X}_i)
\end{equation}
where $\{\theta_{i_k} \in \bm{\theta}_i;k=1,\ldots,K\}$ are the parameters of the Lagrangian, possibly denoting the coefficients of the energy potentials. In a compact matrix, form the above equation can be written as, $\mathcal{L}({\bm{X}_i}, \dot{\bm{X}}_i ) = \mathbf{D}({\bm{X}_i}, \dot{\bm{X}}_i ) \bm{\theta}_i$.
In the remainder of the discussion, we drop the arguments of the $\mathcal{L}(\cdot, \cdot)$ and $\mathbf{D}(\cdot, \cdot)$. To discover the Lagrangian that provides the stationary action, we constrain the Lagrangian to satisfy  \eqref{eq:lagrange_hom} and \eqref{eq:lagrange_nonhom}, whichever is applicable. Since the following procedure is indistinguishable for both the conservative and non-conservative systems, we demonstrate the proposed work using the Lagrangian equation for a conservative system. Thus, we substitute  \eqref{eq:lagrange_lib} in to  \eqref{eq:lagrange_hom}, and obtain,
\begin{equation}\label{eq:lagrange_motion}
    \frac{d}{d t}\left(\frac{\partial \mathbf{D}}{\partial \dot{X}_i}\right) {\bm \theta}_i -\frac{\partial \mathbf{D}}{\partial X_i}{\bm \theta}_i = \bm{0}, \quad i=1,2, \ldots, m .
\end{equation}
% Rearranging the above equation, we obtain,
% \begin{equation}\label{eq:motion_modif}
%     \left[ \frac{d}{d t}\left( \frac{\partial }{\partial \dot{X}_i} \right) - \frac{\partial }{\partial {X}_i} \right] \mathbf{D} \bm{\theta}_i = \bm{0}, \quad i=1,2, \ldots, m .
% \end{equation}
The above equation can be recast in the form of a linear regression problem, which is,
\begin{equation}\label{eq:regression}
    \bm{Y}_i = \hat{\mathbf{D}}\bm{\theta}_i + {\bm{\epsilon}}, \quad i=1,2, \ldots, m ,
\end{equation}
where $\bm{Y}_i \in \mathbb{R}^{N}$ is the zero target vector, $\hat{\mathbf{D}} \in \mathbb{R}^{N \times K}$ is the resulting dictionary matrix when the Euler-Lagrangian operator is evaluated on $\mathbf{D} \in \mathbb{R}^{N \times K}$, and $\bm{\epsilon} \in \mathbb{R}^{N}$ is the model miss-match error. The error $\bm{\epsilon}$ ideally should vanish if the model constructed from the dictionary matches the actual model. Due to the presence of noise in the measurements, the error will not be zero. We seek to find a sparse solution for $\bm{\theta}_i$ by minimizing the error ${\bm{\epsilon}}$. However, when solved, the  \eqref{eq:regression} yields a trivial solution, i.e., $\{\theta_{i_j}, j=1,\ldots,K \}$=0.

\begin{figure}[!ht]
    \centering
    \includegraphics[width=\textwidth]{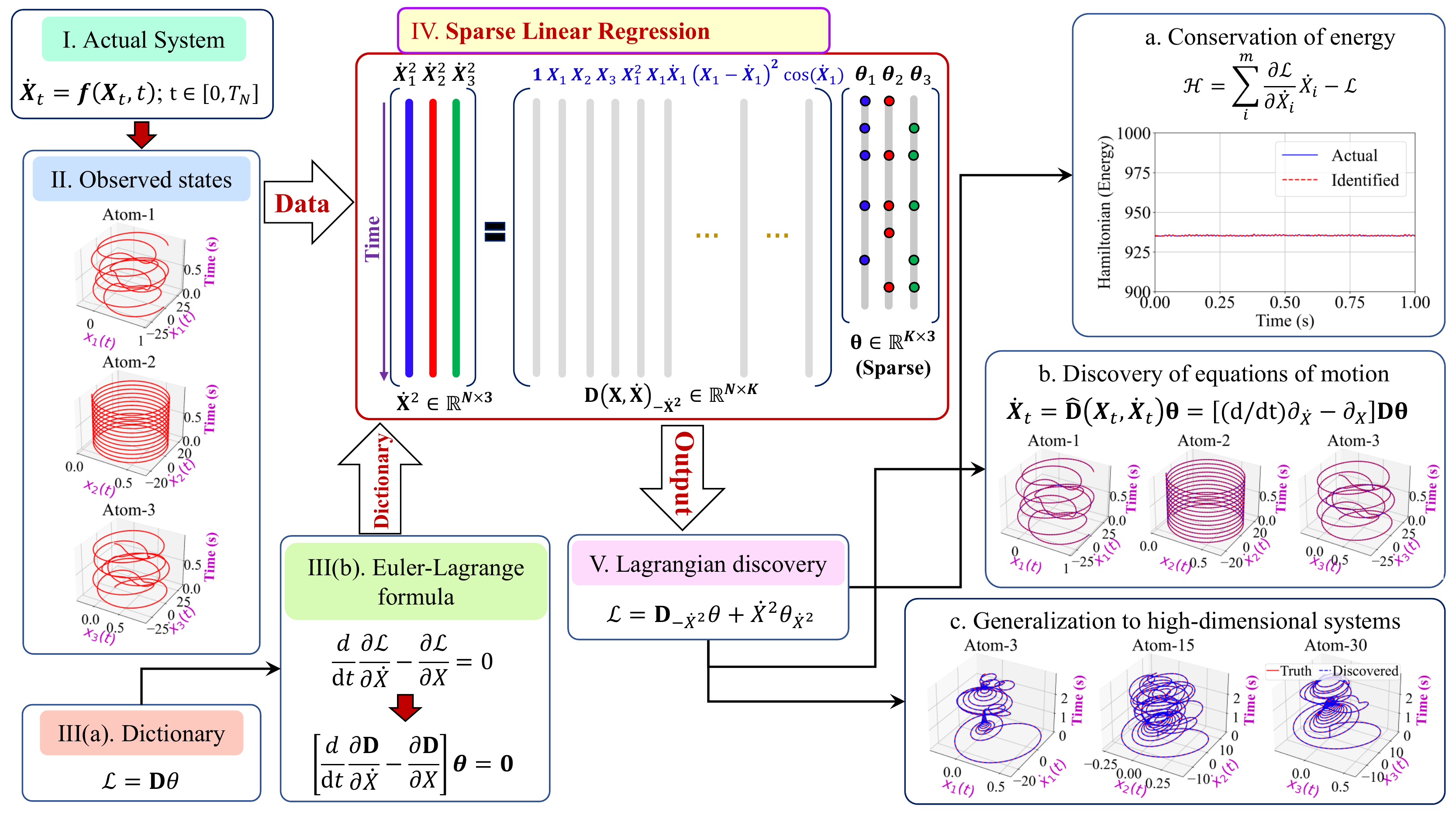}
    \caption{Schematic of the proposed Lagrange discovery algorithm, demonstrated on a triatomic molecule. State observation data (displacement and velocity) are collected from the system. First, the Lagrange is expressed in terms of a dictionary of nonlinear functions of the states. Then the Euler-Lagrange equation is applied to the Library. Next, a target vector is constructed from the column corresponding to the basis $\dot{X}^2$ from the Euler-Lagrange dictionary. A sparse regression is performed between the Euler-Lagrange dictionary and the target vector. Finally, the Lagrangian is discovered by augmenting back the basis $\dot{X}^2$ to the Euler-Lagrange dictionary. The discovered Lagrangian facilitates the direct discovery of Hamiltonian and equations of motion. It also has the potential to generalize to high-dimensional systems, allowing us to learn the desired Lagrangian from a small subset of data.}
    \label{fig:method}
\end{figure}

To obtain the non-trivial solution for the model parameters $\bm{\theta}$, we model the label/target vector of the regression as the vector of $\dot{X}^2$. In particular, we identify the column index of the basis function $\dot{\bm{X}}_i^2$ from the library $\mathbf{D}$ and set the label $\bm{Y}_i$ as the corresponding column from the differentiated library $\hat{\mathbf{D}}$. This results in the following regression equation,
\begin{equation}\label{eq:motion_opt2}
    \hat{\bm{D}}_{\dot{X}_i^2} = \hat{\mathbf{D}}_{(-\dot{X}_i^2)} {\bm{\theta}_i} + {\bm{\epsilon}} , \quad i=1,2, \ldots, m ,
\end{equation}
where $\bm{Y}_i = \hat{\bm{D}}_{\dot{X}_i^2}$ and $\hat{\mathbf{D}}_{(-\dot{X}_i^2)} \in \mathbb{R}^{N \times (K-1)}$ is the $\dot{\bm{X}}_i^2$ removed library. In this way, the proposed candidate library can incorporate all the energy potentials, including kinetic energy functions that are dependent on both displacement and velocities. The above multivariate regression equation can be represented as follows,

\begin{equation}
    \left[ {\begin{array}{*{20}{c}}
    \vert & \vert &  & \vert \\
    \hat{\bm{D}}_{\dot{X}_1^2}& \hat{\bm{D}}_{\dot{X}_2^2}& \ldots &\hat{\bm{D}}_{\dot{X}_m^2}\\
    \vert & \vert &  & \vert 
    \end{array}} \right] = \underbrace {\left[ {\begin{array}{*{20}{c}}
    1 &{{\hat{\ell} _1}({\bm{X}_{1}}) }&{{\hat{\ell} _2}({\dot{\bm{X}}_{1}})}& \cdots &{{\hat{\ell} _K}({\bm{X}_{1}}, {\dot{\bm{X}}_{1}})}\\
    1 &{{\hat{\ell} _1}({\bm{X}_{2}}) }&{{\hat{\ell} _2}({\dot{\bm{X_{2}}}})}& \cdots &{{\hat{\ell} _K}({\bm{X}_{2}}, {\dot{\bm{X}}_{2}})}\\
    \vdots & \vdots & \ddots & \vdots \\
    1 &{{\hat{\ell} _1}({\bm{X}_{N}}) }&{{\hat{\ell} _2}({\dot{\bm{X_{N}}}})}& \cdots &{{\hat{\ell} _K}({\bm{X}_{N}}, {\dot{\bm{X}}_{N}})}
    \end{array}} \right]}_{{\bf{L}} \in {\mathbb{R}^{N \times K}}}\underbrace {\left[ {\begin{array}{*{20}{c}}
    {{\theta _{1_1}}}&{{\theta _{2_1}}}& \cdots &{{\theta _{m_1}}}\\
    {{\theta _{1_2}}}&{{\theta _{2_2}}}& \cdots &{{\theta _{m_2}}}\\
     \vdots & \vdots & \ddots & \vdots \\
    {{\theta _{1_k}}}&{{\theta _{2_k}}}& \cdots &{{\theta _{m_k}}}
    \end{array}} \right]}_{{\bf{\theta }} \in {\mathbb{R}^{K \times m}}}.
\end{equation}
Here,
\begin{equation}
    \hat{\ell}(\cdot) = \left[ \frac{d}{d t}\left( \frac{\partial }{\partial \dot{X}_i} \right) - \frac{\partial }{\partial {X}_i} \right] \ell (\cdot) .
\end{equation}
The degree of accuracy of the discovered Lagrangian entirely depends on the choice and versatility of the dictionary functions. In the absence of prior information, the dictionary may contain a large number of candidate functions. However, from our knowledge, we know for sure that only a few of the candidate potential functions will be active in the final model of Lagrangian potential. This is to say that most of the entries $\theta_k \in \bm{\theta}_i$ will take a zero value. To find the sparse solution vector $\{\theta_{i_1}, \ldots, \theta_{i_K} \}$, we employ the sequential threshold least-square regression proposed in \cite{brunton2016discovering}. 
The sparse algorithm computes a least-squares solution for $\bm{\theta}_i$ and then thresholds solution vector using a sparsity constant $\lambda$. The degree of sparsity is controlled by the constant $\lambda$. This procedure is repeated for several iterations on the remaining non-zero entries of $\bm{\theta}_i$, where the cardinality of non-zero coefficients decreases with each iteration. The procedure is stopped when the non-zero coefficients converge. At each iteration, the sparsity is achieved by penalizing the solution with its $L^1$-norm as follows, 
\begin{equation}\label{eq:sparse_reg}
    {\bm{\theta}}_i = \underset{{\bm{\theta}}_i^{*}}{\arg \mathop{\rm min}} \| \hat{\mathbf{D}}_{(-\dot{X}_i^2)} {\bm{\theta}_i^{*}} - \ddot{\bm{X}}_i \|_2 + \lambda \|{\bm{\theta}}_i^{*} \|_1, \quad i=1,2, \ldots, m ,
\end{equation}
where $\|\cdot \|_1$ and $\|\cdot \|_2$ denote the $L^1$ and $L^2$ norms, respectively. The superscript denotes the solution vector from the previous iteration. Note that it requires a $m$-distinct search to obtain the sparse matrix $\boldsymbol{\theta} = [\bm{\theta}_1, \ldots, \bm{\theta}_m]$, where each column $\bm{\theta}_i$ represents the sparse vector associated with $i^{th}$-DOF. The above sequential threshold least-square regression is a modification of the Least Absolute Shrinkage and Selection Operator (LASSO) \cite{hastie2009elements}. The algorithm is useful for large-scale problems as compared to brute-force combinational alternatives. More details are available in \cite{brunton2016discovering}. It is simple to use and, when used, converges to the sparse solution within a few iterations. More use of the sequential threshold sparse algorithm can be found in biology \cite{mangan2016inferring}, chemistry \cite{hoffmann2019reactive}, fluid mechanics \cite{loiseau2018constrained}, system identification \cite{li2019discovering}, etc. 

We recall that the dictionary ${\mathbf{D}}_{(-\dot{X}_i^2)} \in \mathbb{R}^{N \times (K-1)}$ contains the symbolic functions of potential candidates of the Lagrangian. However, note that the kinetic energy function $\dot{X}_i^2$ is removed from the ${\mathbf{D}} \in \mathbb{R}^{N \times K}$ to obtain the non-trivial solution. Once the coefficient vector $\bm{\theta}_i$ is determined, the exact analytical form of the Lagrangian $\mathcal{L}$ can be obtained from the symbolic library as,
\begin{equation}\label{eq:lagrangian}
    \mathcal{L} \left(\bm{X}_i, \dot{\bm{X}}_i \right) = {\mathbf{D}}_{(-\dot{X}_i^2)} \bm{\theta}_i + \theta_{\dot{X}_i^2} {\dot{X}_i^2},
\end{equation}
where $\theta_{\dot{X}_i^2}$ = 1 and the addition of the second term on the right-hand side is to compensate for the removal of kinetic energy function $\dot{X}_i^2$ from the original dictionary $\mathbf{D}$. The Lagrangian of the $n$-DOF system is then found by summing all the Lagrangians of each element, which is essentially the difference between the total kinetic energy and the total potential energy. For the implementation, follow the Algorithm \ref{algo_wno}. Once the Lagrangian is discovered, the conservation laws can be directly derived using Noether's theorem and Hamiltonian principles.
An additional feature of the proposed framework (see  \eqref{eq:motion_opt2}) directly provides us with the governing motion equation in addition to the Lagrangian. In recent years, an increase in research has been seen on the distillation of the governing physics from data. Since the proposed framework provides a unified framework for discovering both the Lagrangian and governing equations of motion, it may cater to the needs of the discovery of the physics of data. In the results section, we will demonstrate the robustness and application of the proposed framework.

\begin{algorithm}[ht!]
    \caption{Algorithm for discovering Lagrangian from data}\label{algo_wno}
    \begin{algorithmic}[1]
	\Require{Observations $\{ \mathbf{X} \in \mathbb{R}^{N \times m}, \dot{\mathbf{X}} \in \mathbb{R}^{N \times m} \}$, and hyperparameter $\lambda$.}
        \State{Construct a symbolic dictionary $\mathbf{D}(\cdot, \cdot)$.}\Comment{Eq. \eqref{eq:library}}
        \State{Evaluate the dictionary over states $\mathbf{D}(\mathbf{X}, \dot{\mathbf{X}})$.}
        \For{$i$ in $m$}
            \State{Compute $\partial \mathbf{D}/ \partial X_i$.}
            \State{Compute $(d/dt)(\partial \mathbf{D}/ \partial \dot{X}_i)$.}
            \State{Difference $\hat{D} = (d/dt)(\partial \mathbf{D}/ \partial \dot{X}_i - \partial \mathbf{D}/ \partial X_i)$.}
            \State{Locate the column with $\dot{\bm{X}}^2_i$ from the dictionary $\mathbf{D}(\bm{X}_i, \dot{\bm{X}}_i)$.}
            \State{Construct the target vector with $ \bm{Y}_i = \hat{\bm{D}}_{\dot{\bm{X}}^2_i}$.} \Comment{Eq. \eqref{eq:motion_opt2}}
            \State{Perform a sequentially threshold least-squares between $\bm{Y}_i$ and $\hat{\mathbf{D}}_{(-\dot{X}_i^2)}$.} \Comment{Eq. \eqref{eq:sparse_reg}}
            \State{Obtain the Lagrangian $\mathcal{L}(\bm{X}_i, \dot{\bm{X}}_i )$.} \Comment{Eq. \eqref{eq:lagrangian}}
        \EndFor
        \Ensure{Obtain the total Lagrangian $\mathcal{L}(\mathbf{X}, \dot{\mathbf{X}}) = \sum_i \mathcal{L}(\bm{X}_i, \dot{\bm{X}}_i ); i=1,\ldots,m$.}
    \end{algorithmic}
\end{algorithm}

\noindent \textbf{Remark 1} Often, the physical systems of interest are mathematically represented using high-dimensional ordinary (ODE) and partial differential equations (PDE). In the case of ODEs, the data are collected at each DOFs, and in the case of PDEs, the data are observed at discrete spatial locations. In both cases, the proposed framework provides a scalar approach, where the Lagrangian is discovered at each discrete location and then summed to obtain the final Lagrangian density of the system.

\noindent \textbf{Remark 2} Independent of numerical discretization or experimental measurements, if data are collected on a large number of discrete locations (DOFs for the system of ODE and spatial grid of PDEs), then the state dimension-$m$ may be prohibitively large. For example, one can think of the number of molecules in the atomic chain and spatial variables in fluid simulation. Since the cardinality of the dictionary $\mathbf{D}$ increases exponentially with an increase in state dimension-$m$, the proposed framework may become very inefficient. Fortunately, the proposed framework can be applied to a subset of the state dimension to obtain the basic structure of the Lagrangian, which can be generalized to $m$-dimension to obtain the system's Lagrangian density. In the results section, we will provide evidence using a problem on the discovery of Lagrangian for an atomic chain.

\section{Results}\label{sec:numerical}
We showcase the performance of the proposed framework by discovering the Lagrangian of several benchmark problems. Examples include simple linear oscillators, atomic chains, and the movement of transversal waves in solids. We assume that we have access to only a single observation of system states. For the discovery of the Lagrangian of the forced oscillator, it is assumed that both input-output measurements are available. Here, input means the excitation, and output means the state observations. The parameters used during the simulation are provided in Table \ref{table_param}. In all the examples, the measurements are arranged in a matrix form (i.e., $\mathbf{X}$, $\dot{\mathbf{X}}$) where the row represents time-snapshots of the dynamics, and the columns represent the degrees of freedom. Then each column of the dictionary (see  \eqref{eq:library}) is evaluated on the measurement matrices. Reiterating, we want to discover a sparse representation of the Lagrangian. We utilize the sparse regression in  \eqref{eq:sparse_reg} to discover the active basis function from the dictionary. Later, we employ the Hamiltonian principle to discover the underlying conservation laws from the discovered Lagrangian. For each example, we further compare the discovered governing equations with their actual counterparts. Towards the end of this section, we explore additional features of the proposed framework, such as zero-shot generalization, perpetual predictive capability, and generalization to high-dimensional systems. 

\begin{table}[ht]
    \centering
    \caption{Simulation parameters of the example problems.}
    \label{table_param}
    \begin{threeparttable}
	\begin{tabular}{p{8cm}l}
		\toprule
		\textbf{Systems} & \textbf{Parameters} \\
		\midrule
		Harmonic oscillator$^{ (\ref{example_harmonic})}$ & $m$=10kg, $k$=5000N/m$^2$, $A$=10 \\
		% \hdashline
		Pendulum$^{ (\ref{example_3})}$ & $m$=1kg, $l$=2m, $g$=9.81m/s$^2$ \\
		% \hdashline
		Undamped MDOF vibration$^{ (\ref{example_4})}$ & $m_i$ = 10, $k_i$ = 5000; $i=1,2,3$ \\
		% \hdashline
		Vibration of a Linear Triatomic Molecule$^{ (\ref{example_5})}$ & $m_{1,2}$=1AMU, $m_2$=2AMU$^{\dagger}$, $k_i$=1870N/m$^2$; $i=1,2,3$ \\
		% \hdashline
		Elastic Transversal Waves in a Solid$^{ (\ref{example_6})}$ & $c$=25 ($\rho=960kg/m^3$, $\delta$=0.01, $\mu$=4GPa) \\
		% \hdashline
		Flexion Vibration of a Blade$^{ (\ref{example_7})}$ & $c$=1 ($\rho$=1, $E$=1, $S$=1, $I$=1, $\delta$=0.01) \\
		\midrule
	\end{tabular}
    \begin{tablenotes}
        \item \textbf{Note}: $\dagger$ AMU = atomic mass unit. For all the demonstrations in this work, a single set of system states are simulated using the fourth-order Runge-Kutta scheme. The noise in the measurements is modeled as $N$-dimensional sequence of zero-mean Gaussian white noise with a standard deviation equal to 5\% of the standard deviation of the simulated quantities.
    \end{tablenotes}
    \end{threeparttable}
\end{table}

\subsection{Harmonic oscillator}\label{example_harmonic}
As a first example, we consider the undamped vibration of a mass attached to a spring. We consider both the free and forced vibration of the harmonic oscillator. The corresponding governing equations for the free and forced vibration cases are as follows,
\begin{subequations}
    \begin{equation}\label{eq:free}
        m\ddot{X}(t) + kX(t) = 0; \quad X(t=0)=X_0, \quad t \in [0,T]
    \end{equation}
    \begin{equation}\label{eq:forced}
        m\ddot{X}(t) + kX(t) = F(t); \quad X(t=0)=X_0, \; F(t)>0, \quad t \in [0,T]
    \end{equation}
\end{subequations}
where $m$ and $k$ are the mass and associated spring stiffness, respectively. For the forced vibration equation, the excitation is modeled as $F(t) = A{\rm{sin}}(2 \pi t)$. The associated Lagrangian for the free and forced vibration cases are $\mathcal{L}$=$\frac{1}{2}m\dot{X}^2 - \frac{1}{2}kX^2$, and $\mathcal{L}$=$\frac{1}{2}m\dot{X}^2 - \frac{1}{2}kX^2 + XF(t)$, respectively. An initial condition of $\{X, \dot{X}\}_{(t=0)}$=\{1,0\} is used to simulate the data. The sampling frequency and total duration of time $T$ are taken as \{1000Hz, 1s\} for  \eqref{eq:free} and \{2000Hz, 2s\} for  \eqref{eq:forced}. We construct the differentiated library $\hat{\mathbf{D}}$ and then obtain the regression equation in  \eqref{eq:motion_opt2} by constructing the target vector as the column of $\dot{X}^2$ from $\hat{\mathbf{D}}$. Finally,  \eqref{eq:lagrangian} is used to obtain the underlying Lagrangian in mass-normalized form. 
\textbf{Results}: The discovery results are summarized in Fig. \ref{fig:spy_linear} and \ref{fig:spy_forced}. As depicted in Fig. \ref{fig:spy_linear} and \ref{fig:spy_forced}, it can be concluded that the proposed framework is able to distill the correct basis functions for the harmonic oscillator. The identified parameter of the potential energy is given in Table \ref{tab:ident_param}, from which we can also ascertain that the proposed framework can discover the near-exact system parameters.

\begin{table}[ht]
    \centering
    \caption{Identification of the parameters of the system}
    \begin{tabular}{lllll}
        \toprule
        & Parameters & Actual values & Identified values & Relative Error (\%) \\
        \midrule
        Ex. (\ref{example_harmonic}\eqref{eq:free}) & k/m & 500 & 500.14 & 0.028 \\
        Ex. (\ref{example_harmonic}\eqref{eq:forced}) & \{k/m, F/m\} & \{500,1\} & \{499.99, 1.01\} & \{0.198, 1\} \\
        Ex. (\ref{example_3}) & g/l & 9.81 & 9.82 & 0.101 \\
        Ex. (\ref{example_4}) & $k_i$/$m_i$ & 500 & 499.87 & 0.026 \\
        Ex. (\ref{example_5}) & $k_i$/$m_i$ & 1870 & 1870.67 & 0.036 \\
        Ex. (\ref{example_6}) & $c$ & 25 & 24.98 & 0.08 \\
        Ex. (\ref{example_7}) & $c$ & 1$\times 10^{8}$ & 1.018$\times 10^{8}$ & 1.81 \\
        \midrule
    \end{tabular}
    \label{tab:ident_param}
\end{table}

\begin{table}[ht]
	\centering
	\caption{Lagrangian of the undertaken examples.}
	\label{table_lagrange}
	\begin{tabular}{p{2.5cm}p{4cm}l}
		\hline
		\textbf{Systems} & \textbf{Motion equation} & \textbf{Lagrangian} \\
		\hline
		Ex. (\ref{example_harmonic}\eqref{eq:free}) & $m\ddot{X}(t) + kX(t) = 0$ & $\mathcal{L}$=$\frac{1}{2}m\dot{X}^2 - \frac{1}{2}kX^2$ \\
		Ex. (\ref{example_harmonic}\eqref{eq:forced}) & $m\ddot{X}(t) + kX(t) = F(t)$ & $\mathcal{L}$=$\frac{1}{2}m\dot{X}^2 - \frac{1}{2}kX^2 + XF(t)$ \\
		Ex. (\ref{example_3}) & $m l^2 \ddot{\theta}(t) + mgl {\rm{sin}} \theta(t) = 0$ & $\mathcal{L}$=$\frac{1}{2}m{l^2}\dot{\theta}^2 - mgl{\rm{cos}}\theta$ \\
		Ex. (\ref{example_4}) & $\mathbf{M}\ddot{\bm{X}}(t) + \mathbf{K} \bm{X}(t) = 0$ & $\mathcal{L}$=$\frac{1}{2} \sum_{k=1}^{3} m_k \dot{X}_k^2 - \left( \frac{1}{2}k_1 X_1^2 + \frac{1}{2}k_2(X_2-X_1)^2 + \frac{1}{2}k_3(X_3-X_2)^2 \right)$ \\
		Ex. (\ref{example_5}) & $\mathbf{M}\ddot{\bm{X}}(t) + \mathbf{K} \bm{X}(t) = 0$ & $\mathcal{L}$=$\frac{1}{2} \sum_{k=1}^{3} m_k \dot{X}_k^2 - \left( \frac{1}{2}k_2(X_2-X_1)^2 + \frac{1}{2}k_3(X_3-X_2)^2 \right)$ \\
		Ex. (\ref{example_6}) & ${\partial^2_t u(X,t)} = c^2 {\partial^2_x u(X,t)}$ & $\mathcal{L}$=$\frac{1}{2}\sum_{i} \rho S \delta \dot{u}_i^2 - \sum_i {\mu S}{\delta}^{-1} \left(u_{i+1}-u_{i} \right)^2$ \\
		Ex. (\ref{example_7}) & ${\partial^2_t u(X,t)} = c^2 {\partial^4_x u(X,t)}$ & $\mathcal{L}$=$\frac{1}{2} \mu \delta \sum_{i} \dot{u}_i^2 - {EI}({2 \delta^3})^{-1} \sum_i \left(u_{i+1}-2u_{i}+u_{i-1} \right)^2$ \\
		\hline
	\end{tabular}
\end{table}

\subsection{Simple pendulum}\label{example_3}
Next, we consider the motion of a simple pendulum with a mass $m$ under the action of gravity. The purpose of this example is to demonstrate the performance of the proposed framework for the discovery of the Lagrangian from angular measurements. The governing equation of motion of a simple pendulum is given as follows,
\begin{equation}
    m l^2 \ddot{\theta}(t) + mgl {\rm{sin}} \theta(t) = 0; \quad \theta(t=0)=\theta_0, \quad t \in [0,T]
\end{equation}
where $l$ is the length of the suspension. For the simulation of the data, we use a sampling frequency of 1000Hz and a total duration of $T$=10s. The initial angular displacement and velocity are considered as $\{\theta, \dot{\theta}\}_{(t=0)}$=\{0.5,0\}. The associated Lagrangian is $\mathcal{L}$=$\frac{1}{2}m{l^2}\dot{\theta}^2 - mgl{\rm{cos}}\theta$. Here again, we have used the column $\dot{\theta}^2$ from the library $\hat{\mathbf{D}}$ as the target vector for the regression problem in  \eqref{eq:motion_opt2} to discover the underlying Lagrangian successfully. 
\textbf{Results}: The results are summarized in Fig. \ref{fig:spy_pendulam}. From the figure, we see that the proposed framework has correctly identified the kinetic energy function $\dot{\theta}^2$ and the harmonic function $cos(\cdot)$. The associated parameters are provided in Table \ref{tab:ident_param}. Note that the parameters are discovered in their mass-normalized form.

\subsection{Undamped MDOF vibration}\label{example_4}
\begin{figure}[th!]
     \centering
     \begin{subfigure}[b]{0.75\textwidth}
        \centering
        \includegraphics[width=\textwidth]{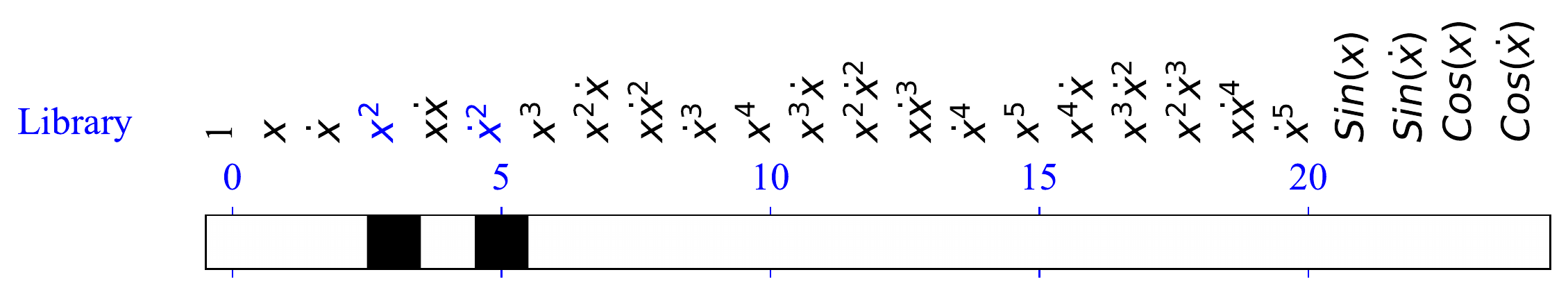}
        \caption{Harmonic oscillator}
        \label{fig:spy_linear}
     \end{subfigure}
    %  \hfill
     \begin{subfigure}[b]{0.75\textwidth}
        \centering
        \includegraphics[width=\textwidth]{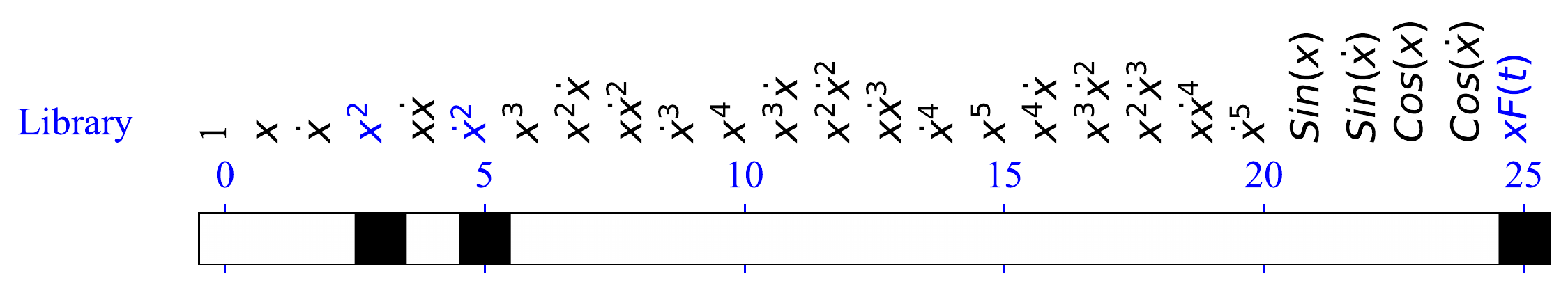}
        \caption{Forced oscillator}
        \label{fig:spy_forced}
     \end{subfigure}
    %  \hfill
     \begin{subfigure}[b]{0.75\textwidth}
        \centering
        \includegraphics[width=\textwidth]{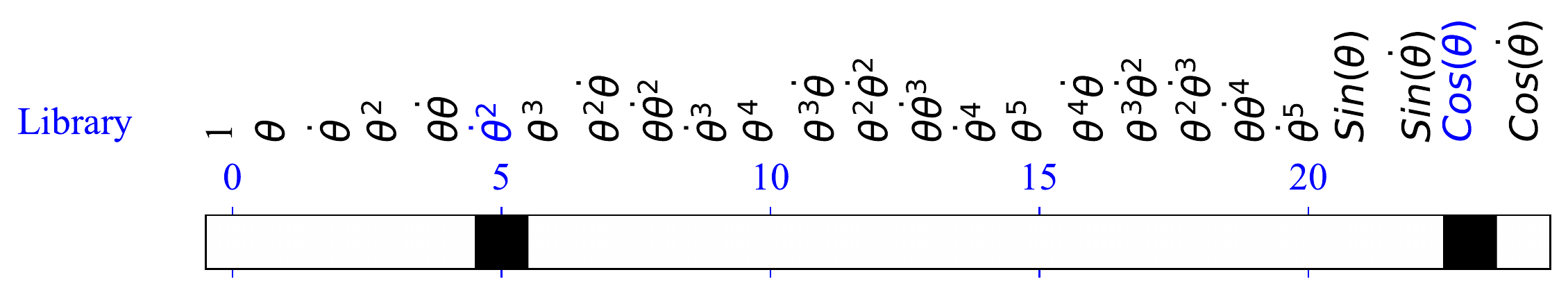}
        \caption{Pendulum}
        \label{fig:spy_pendulam}
     \end{subfigure}
    %  \hfill
     
     \begin{subfigure}[b]{0.4\textwidth}
        \centering
        \includegraphics[width=\textwidth]{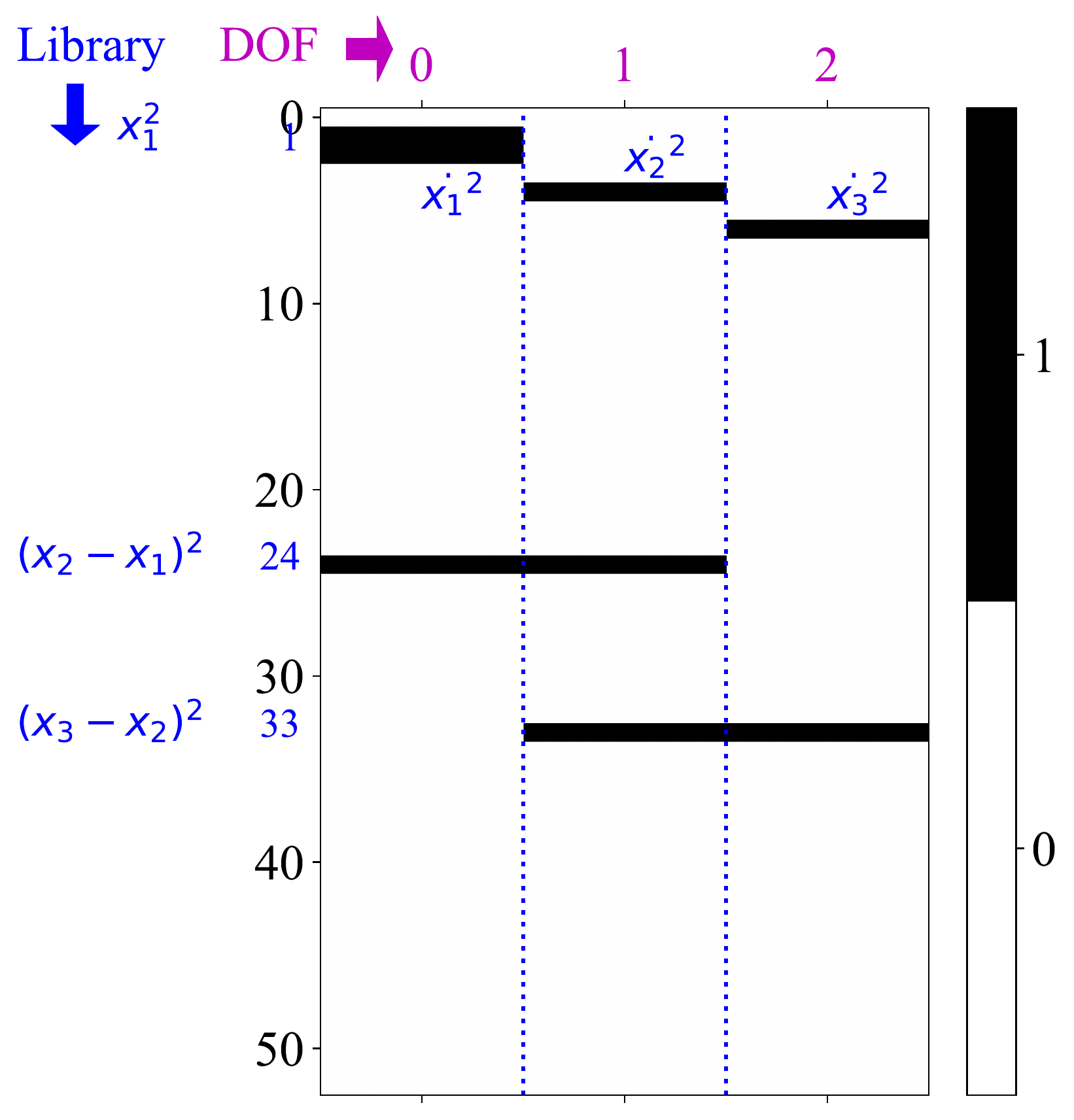}
        \caption{3DOF dynamical system}
        \label{fig:spy_3dof}
     \end{subfigure}
    %  \hfill
     \begin{subfigure}[b]{0.4\textwidth}
        \centering
        \includegraphics[width=\textwidth]{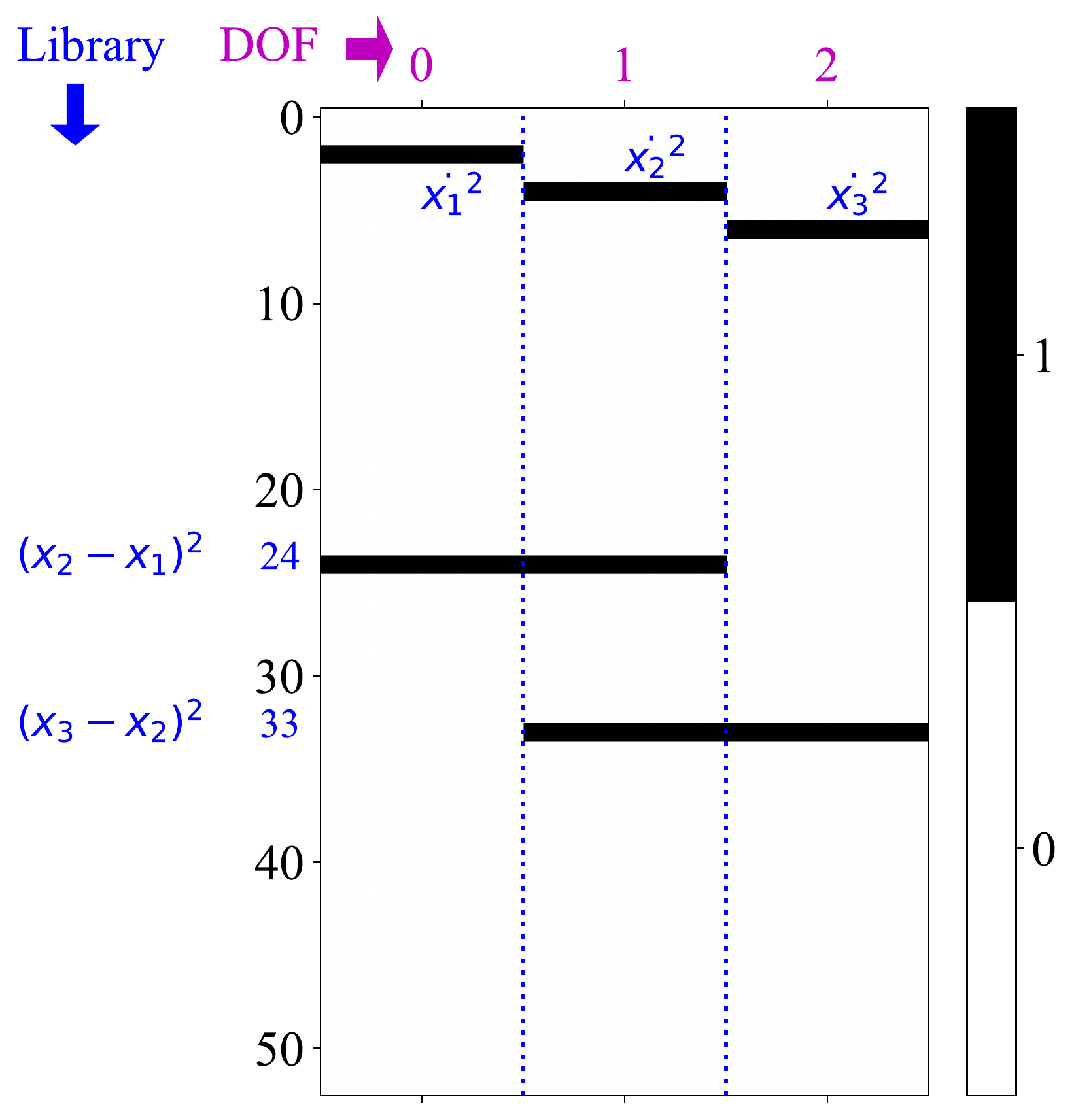}
        \caption{Triatomic molecular chain}
        \label{fig:spy_atom}
     \end{subfigure}
        \caption{\textbf{Identification of the active basis functions from the library}. The identified basis functions are shown. (a) Harmonic oscillator: the potential and the kinetic energy basis functions $X^2$ and $\dot{X}^2$ are found to be active from $\mathbf{D}\in \mathbb{R}^{25}$. (b) Forced oscillator: the potential, the kinetic and forced potential functions $X^2$ and $\dot{X}^2$ and $XF(t)$ are found to be active $\mathbf{D}\in \mathbb{R}^{26}$. (c) Pendulum: the Lagrangian bases $\dot{\theta}^2$ and $\cos(X)$ is found to be active $\mathbf{D}\in \mathbb{R}^{25}$. (d) 3DOF dynamical system: the library has $\mathbf{D}\in \mathbb{R}^{52}$ 52 basis functions. For the first DOF, the bases $X_1^2$, $\dot{X}_1^2$, and $(X_2-X_1)^2$; for the second DOF, the bases $\dot{X}_2^2$, $(X_2-X_1)^2$, and $(X_3-X_2)^2$, and for the third DOF the bases $\dot{X}_3^2$ and $(X_3-X_2)^2$ are identified to be active. (e) Triatomic molecular chain: for the first DOF, the basis functions $\dot{X}_1^2$ and $(X_2-X_1)^2$, for the second DOF, the basis functions $\dot{X}_2^2$, $(X_2-X_1)^2$ and $(X_3-X_2)^2$, and for the third DOF, the bases $\dot{X}_3^2$ and $(X_3-X_2)^2$ are active from $\mathbf{D}\in \mathbb{R}^{52}$.}
        \label{fig:spy_ode}
\end{figure}

We consider an oscillator with three degrees of freedom in this example to demonstrate the applicability of the proposed framework for systems represented using a system of ODEs. Let the mass and stiffness of the three DOF systems be denoted as $m_i$ and $k_i$ for $i=1,\ldots,3$. The equations of motion for this 3DOF can be derived as follows,
\begin{equation}\label{eq:mdof}
    \mathbf{M} \ddot{\bm{X}}(t) + \mathbf{K} \bm{X}(t) = 0
\end{equation}
where 
\begin{equation}
    \mathbf{M} = \left[ { \begin{array}{ccc}
         m_1 & 0 & 0 \\
         0 & m_2 & 0 \\
         0 & 0 & m_3
    \end{array} } \right], \; \text{and,} \;
    \mathbf{K} = \left[ { \begin{array}{ccc}
         k_1+k_2 & -k_2 & 0 \\
         -k_2 & k_2+k_3 & -k_3 \\
         0 & -k_3 & k_3
    \end{array} } \right]
\end{equation}
Similar to previous examples, we aim to discover the mass-normalized Lagrangian for this example. However, instead of directly discovering the Lagrangian of the whole system, we discover the Lagrangian of each DOF and then sum over all the DOF to obtain the Lagrangian of the complete system. For data simulation, we use the initial conditions $\bm{X} = \{1,0,0,0,0,0\}$. A sampling frequency of 1000Hz and $T$=1s is used. The actual Lagrangian for this systems is $\mathcal{L}$=$\frac{1}{2} \sum_{k=1}^{3} m_k \dot{X}_k^2 - \left( \frac{1}{2}k_1 X_1^2 + \frac{1}{2}k_2(X_2-X_1)^2 + \frac{1}{2}k_3(X_3-X_2)^2 \right)$. To discover the Lagrangian of each DOF, we use a similar procedure as mentioned in previous examples. 
\textbf{Results}: The results are summarized in Fig. \ref{fig:spy_3dof}. In Fig. \ref{fig:spy_3dof}, the active dictionary functions for each DOF are portrayed. A value of 1 (black color) means the dictionary function is active, and a value of 0 (white color) means the dictionary function is not active. The true Lagrangian for the 3DOF system is discovered by a scalar approach, where Lagrangian for each DOF is discovered and then summed to obtain the system's Lagrangian. In the first DOF, the bases $X_1^2$ and $(X_2-X_1)^2$, for the second DOF, the bases $(X_2-X_1)^2$ and $(X_3-X_2)^2$, and for the third DOF the basis $(X_3-X_2)^2$ are found to active. When the Lagrangian obtained from each DOFs is summed, we obtain the actual Lagrangian for the system.

\subsection{Vibration of a Triatomic Molecule}\label{example_5}
In this example, we are interested in learning the Lagrangian associated with longitudinal vibrations of a triatomic molecule. The molecule is composed of two diﬀerent types of atoms, disposed of in an alternate way. We employ a simple linear and symmetric model for the triatomic chain with respective point-like masses $m-\tilde{m}-m$, where $\tilde{m}$ denotes the mass of the heavy atom and $m$ denotes the mass of light ones. The mass of the light atom is expressed in terms of the heavy atom as $\tilde{m} = rm$, with $r<1$. The attractive forces between these masses are modeled by two identical springs of constant $k$. We further assume that the deviation from the equilibrium position is small; thus, the harmonic approximation remains valid during the interaction. The interaction between the atoms is governed by  \eqref{eq:mdof}, where
\begin{equation}
    \mathbf{M} = m\left[ { \begin{array}{ccc}
         1 & 0 & 0 \\
         0 & r & 0 \\
         0 & 0 & 1
    \end{array} } \right], \; \text{and,} \;
    \mathbf{K} = k\left[ { \begin{array}{ccc}
         1 & -2 & 0 \\
         2 & 2 & -3 \\
         0 & -2 & 3
    \end{array} } \right]
\end{equation}
where $r=M/m$. The data are obtained for $T$=1s using a sampling rate 1000Hz. An initial condition of $\bm{X}=[1,0,0,0,0,0]$ is used. The parameters for the triatomic chain are given in Table \ref{table_param}. Similar to the previous MDOF example, we aim to discover the Lagrangian for the triatomic chain by discovering the individual Lagrangian of each molecule and then summing up them. For each molecule, we use the procedure described in  \eqref{eq:motion_opt2} and \eqref{eq:lagrangian}. The actual Lagrangian of the triatomic molecule is given as $\mathcal{L}$=$\frac{1}{2} \left( \sum_{k=1}^{2} m_k \dot{X}_k^2 + \tilde{m} \dot{X}_{\tilde{m}}^3\right) - \left( \frac{1}{2}k_2(X_2-X_1)^2 + \frac{1}{2}k_3(X_3-X_2)^2 \right)$.
\textbf{Results}: The results ate summarized in Fig. \ref{fig:spy_atom}. In the figure, we see that the discovered Lagrangian for the first DOF is a function of the basis $(X_2-X_1)^2$. For the second DOF, the bases are $(X_2-X_1)^2$ and $(X_3-X_2)^2$, and for the third DOF, the basis is $(X_3-X_2)^2$. By performing a weighted sum (the parameters are given in Table \ref{tab:ident_param}) of the basis functions, we eventually obtain the actual Lagrangian of the triatomic molecule. 

\begin{figure}[ht!]
     \centering
     \begin{subfigure}[b]{0.4\textwidth}
        \centering
       \includegraphics[width=\textwidth]{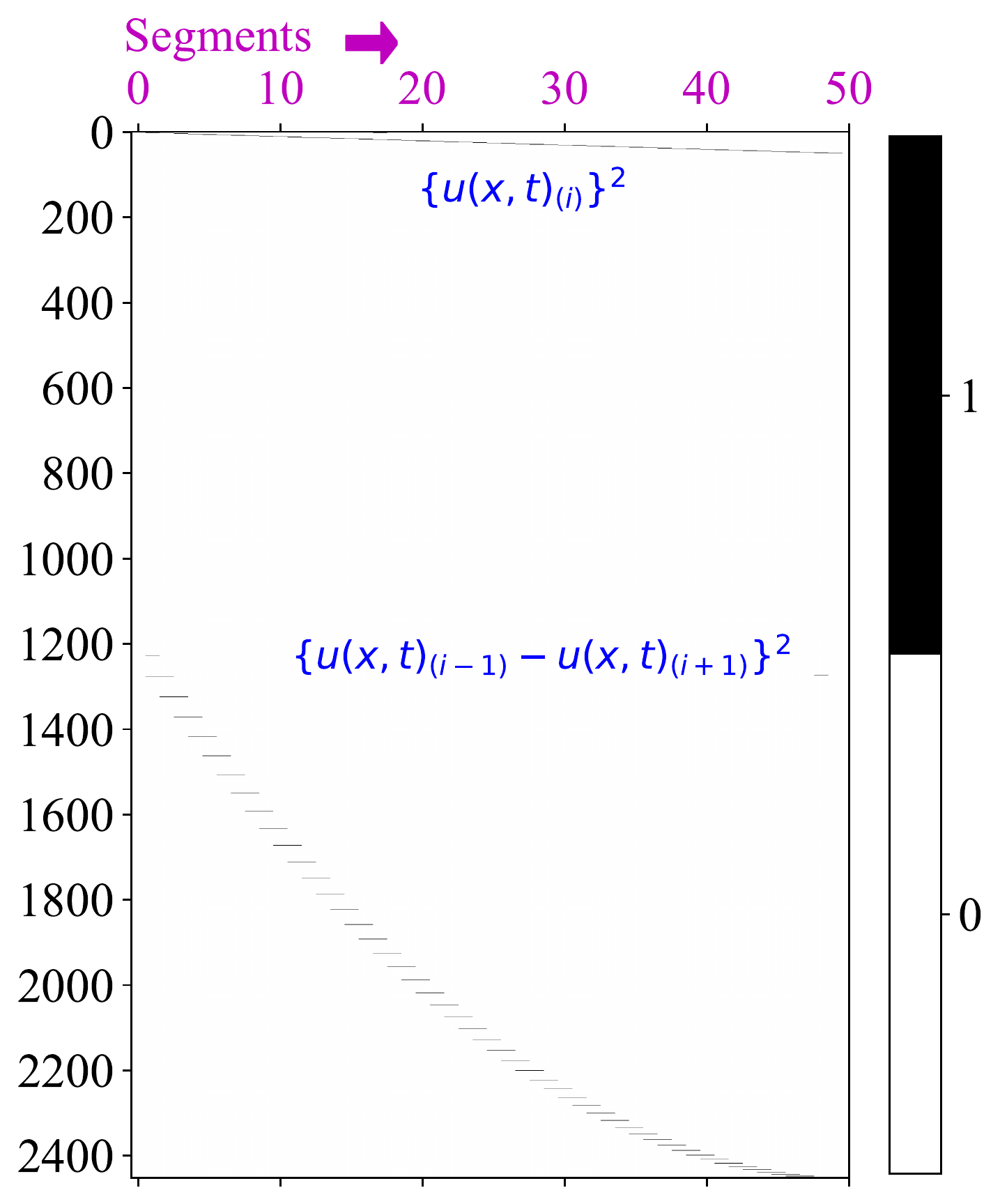}
        \caption{Elastic Transversal Waves in a Solid}
        \label{fig:spy_string}
     \end{subfigure}
     \begin{subfigure}[b]{0.45\textwidth}
        \centering
        \includegraphics[width=\textwidth]{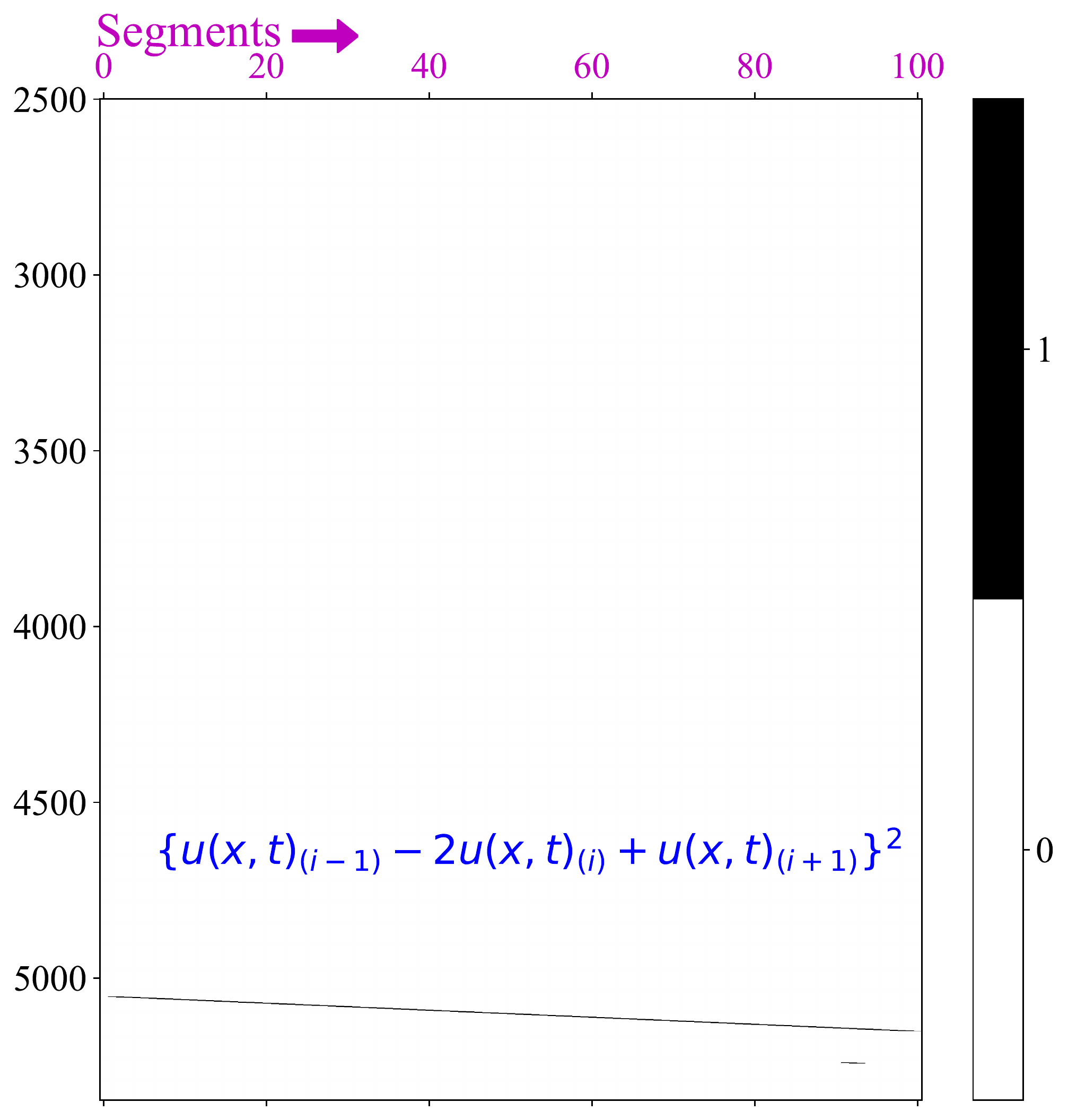}
        \caption{Flexion Vibration of a Blade}
        \label{fig:spy_beam}
     \end{subfigure}
        \caption{\textbf{Identification of the kinetic energy bases from the library for PDEs}. The weights of the dictionary functions are portrayed. (a) Elastic Transversal Waves in a Solid: for the $i^{th}$-discretization, the kinetic energy bases are found as $\left(u_{i+1}-u_{i} \right)^2$. (b) Flexion Vibration of a Blade: for the $i^{th}$-discretization, the kinetic energy bases $\left(u_{i+1}-2u_{i}+u_{i-1} \right)^2$ are found to be active.}
        \label{fig:spy_pde}
\end{figure}

\subsection{Elastic Transversal Waves in a Solid}\label{example_6}
In this example, we extend the proposed framework for discovering Lagrangian for PDE problems. We consider a string with section $S$ and with mass density $\rho$ that is deformed transversely. This example can also be related to the transmission of longitudinal waves (often known as S waves) in fluids. We aim to discover the Lagrangian that will govern the time evolution of deformation in the string. The motion in the string is described by the following partial differential equation,
\begin{equation}
    \frac{\partial^2 u(X,t)}{\partial t^2} = c^2 \frac{\partial^2 u(X,t)}{\partial X^2}
\end{equation}
where $c=\sqrt{\mu / \rho}$ is the speed of elastic transversal wave, $\mu$ is the shearing modulus, and $\rho$ is the mass density of the material. It can be noticed that the governing equation is exactly the one-dimensional wave equation. Let the length of the string be $l$. For data simulation, we virtually cut the string length $l$ into $N+1$ slices with thickness $\delta$. We refer to each slice as a node and place an identical mass at the center of each slice. During deformation, these nodes move transversely by an amount of $u_i$ with respect to an equilibrium condition. The total kinetic $T$ and potential energy $V$ of the system are given as,
\begin{equation}
    T = \frac{1}{2} \rho S \delta \sum_i \dot{u}_i^2, \; V = \frac{\mu S}{\delta} \sum_i \left(u_{i+1}-u_{i} \right)^2
\end{equation}
The associated Lagrangian is given by $\mathcal{L}$=$\frac{1}{2}\sum_{i} \rho S \delta \dot{u}_i^2 - \sum_i {\mu S}{\delta}^{-1} \left(u_{i+1}-u_{i} \right)^2$. More details are given in \cite{gignoux2009solved}. For data simulation, an initial profile of ${\rm{cos}}(2\pi l)$ is applied to the string. A finite difference code is written for the simulation of the data. The parameters for the simulation are provided in Table \ref{table_param}. To discover the Lagrangian, we treat the problem as a high-dimensional extension of the previous MDOF examples. The Lagrangian at discretized spatial locations is discovered and then summed to obtain the Lagrangian of waves. A 50-dimensional spatial grid is created to balance the accuracy and computational time. To simulate the data, we used $\delta$=0.01, $\Delta t$=0.0001s, $T$=1s.
\textbf{Results}: The weights of the dictionary functions are shown in Fig. \ref{fig:spy_string}, where a value of 1 means the dictionary function is active in discovered Lagrangian, and a value of 0 means not-active. Similar to previous examples, the results are shown for the identification of the Lagrangian functions. The active energy functions are also displayed in the figure. It can be seen that out of $\mathbf{R}^{2451}$-basis functions, only one basis is active at each discretized spatial location. This indicates the highly sparse nature of the identified Lagrangian model. By summing the Lagrangian at each spatial location using the parameters in Table \ref{tab:ident_param}, we obtain the actual Lagrangian.

\begin{table}[ht]
    \centering
  \begin{threeparttable}
	\centering
	\caption{Discovery of Lagrangian and Energy conservation (Hamiltonian) in the example problems}
	\label{table:identify}
	\begin{tabular}{p{2.5cm}p{10cm}ll}
		\toprule
		\textbf{Systems} & \textbf{Lagrangian/Hamiltonian} & \textbf{Discovery} & \textbf{$^{\dagger}$Error} (\%) \\
		\midrule
		Ex. (\ref{example_harmonic}\eqref{eq:free}) &  $\mathcal{L}$=$\frac{1}{2}\dot{X}^2 - \frac{1}{2}({k}/{m})X^2$ & \textcolor{blue}{Yes} & $\approx$ 0.0146 \\
		&  $\mathcal{H}$=$\frac{1}{2}\dot{X}^2 + \frac{1}{2}({k}/{m})X^2$ & \textcolor{blue}{Yes} & $\approx$ 0.0986 \\
		\hdashline
		Ex. (\ref{example_harmonic}\eqref{eq:forced}) &  $\mathcal{L}$=$\frac{1}{2}\dot{X}^2 - \frac{1}{2}({k}/{m})X^2 + X({F(t)}/{m})$ & \textcolor{blue}{Yes} & $\approx$ 0.0028 \\
		&  $\mathcal{H}$=$\frac{1}{2}\dot{X}^2 + \frac{1}{2}({k}/{m})X^2 + X({F(t)}/{m})$ & \textcolor{blue}{Yes} & $\approx$ 0.0006 \\
		\hdashline
		Ex. (\ref{example_3}) &  $\mathcal{L}$=$\frac{1}{2}\dot{\theta}^2 - ({g}/{l}){\rm{cos}}\theta$ & \textcolor{blue}{Yes} & $\approx$ 0.0683 \\
		&  $\mathcal{H}$=$\frac{1}{2}\dot{\theta}^2 + ({g}/{l}){\rm{cos}}\theta$ & \textcolor{blue}{Yes} & $\approx$ 0.0776 \\
		\hdashline
		Ex. (\ref{example_4}) &  $\mathcal{L}$= $\frac{1}{2} \sum_{k=1}^{3} m_k \dot{X}_k^2 - \left( \frac{1}{2}k_1 X_1^2 + \frac{1}{2}k_2(X_2-X_1)^2 + \frac{1}{2}k_3(X_3-X_2)^2 \right)$ & \textcolor{blue}{Yes} & $\approx$ 0.1554 \\
		&  $\mathcal{H}$= $\frac{1}{2} \sum_{k=1}^{3} m_k \dot{X}_k^2 + \left( \frac{1}{2}k_1 x_1^2 + \frac{1}{2}k_2(X_2-X_1)^2 + \frac{1}{2}k_3(X_3-X_2)^2 \right)$ & \textcolor{blue}{Yes} & $\approx$ 0.0101 \\
		\hdashline
		Ex. (\ref{example_5}) &  $\mathcal{L}$= $\frac{1}{2} \sum_{k=1}^{3} m_k \dot{X}_k^2 - \left( \frac{1}{2}k_2(X_2-X_1)^2 + \frac{1}{2}k_3(X_3-X_2)^2 \right)$ & \textcolor{blue}{Yes} & $\approx$ 0.3916 \\
		&  $\mathcal{H}$= $\frac{1}{2} \sum_{k=1}^{3} m_k \dot{X}_k^2 + \left( \frac{1}{2}k_2(X_2-X_1)^2 + \frac{1}{2}k_3(X_3-X_2)^2 \right)$ & \textcolor{blue}{Yes} & $\approx$ 0.0127 \\
		\hdashline
		Ex. (\ref{example_6}) &  $\mathcal{L}$=$\frac{1}{2}\sum_{i} \rho S \delta \dot{u}_i^2 - \sum_i ({\mu S}/{\delta}) \left(u_{i+1}-u_{i} \right)^2$ & \textcolor{blue}{Yes} & $\approx$ 1.0092 \\
		&  $\mathcal{H}$=$\frac{1}{2}\sum_{i} \rho S \delta \dot{u}_i^2 + \sum_i ({\mu S}/{\delta}) \left(u_{i+1}-u_{i} \right)^2$ & \textcolor{blue}{Yes} & $\approx$ 0.0645 \\
		\hdashline
		Ex. (\ref{example_7}) &  $\mathcal{L}$=$\frac{1}{2} \mu \delta \sum_{i} \dot{u}_i^2 - ({EI}/{2 \delta^3}) \sum_i \left(u_{i+1}-2u_{i}+u_{i-1} \right)^2$ & \textcolor{blue}{Yes} & $\approx$ 4.9482 \\
		&  $\mathcal{H}$=$\frac{1}{2} \mu \delta \sum_{i} \dot{u}_i^2 + ({EI}/{2 \delta^3}) \sum_i \left(u_{i+1}-2u_{i}+u_{i-1} \right)^2$ & \textcolor{blue}{Yes} & $\approx$ 1.8730 \\
		\bottomrule
	\end{tabular}
	\begin{tablenotes}
      \item $^{\dagger}$Here, the error denotes the mean absolute relative error between the actual and discovered parameter.
    \end{tablenotes}
  \end{threeparttable}
\end{table}

\subsection{Flexion Vibration of a Blade}\label{example_7}
In the last example, we consider a homogeneous, uniform, thin blade having the mass $M$, length $l$, and linear mass $\mu$. At rest, the blade lies in a horizontal position. When this blade is subjected to deformation, it attempts to return to its equilibrium state by transforming its elastic potential energy into kinetic energy. This phenomenon causes vibration in the blade. We intend to discover the Lagrangian (by neglecting the effect of gravity) associated with the vibration of this blade. The motion equation for the vibration in the blade is given by the following differential equation,
\begin{equation}
    \frac{\partial^2 u(X,t)}{\partial t^2} = c^2 \frac{\partial^4 u(X,t)}{\partial X^2}
\end{equation}
with $c= {EI}/{\mu}$. Here, $E$ is the modulus of elasticity of blade material, and $I$ is the moment-of-inertia. For data simulation, we first constructed the discontinuous model of the blade by discretizing the blade length $l$ into $N$ segments of length $\delta$ and mass $\mu \delta$. We refer to these segments as nodes. In the field, these nodes can be related to the sensor locations. We assume the absence of torsion, i.e., only the blade motion in the vertical plane, is considered. Further, we assume that the deformation is weak, which allows making the hypothesis that each segment has a ﬁxed length $\delta$. Under these assumptions the total kinetic ($T$) and potential energy ($V$) are given as,
\begin{equation}
    T = \frac{1}{2} \mu \delta \sum_{i} \dot{u}_i^2, \; V = \frac{EI}{2 \delta^3}\left(u_{i+1}-2u_{i}+u_{i-1}\right)^2
\end{equation}
where $\mu = M/l$ is the mass per unit length. Here, $u_{i}$, $u_{i+1}$, and $u_{i-1}$ denote the deviations at the center of mass, right end, and left end of the $i^{th}$ node from the equilibrium point, respectively. The corresponding Lagrangian is 
$\mathcal{L}$=$\frac{1}{2} \mu \delta \sum_{i} \dot{u}_i^2 - {EI}({2 \delta^3})^{-1} \sum_i \left(u_{i+1}-2u_{i}+u_{i-1} \right)^2$. For more information, readers are referred to \cite{gignoux2009solved}. To obtain the data, the blade is excited using its fundamental frequency (i.e., the first frequency of a cantilever beam). The data are obtained using the finite element method. The simulation parameters are given in Table \ref{table_param}. The Spatio-temporal parameters are $\delta$=0.01, $\Delta t$=0.001s, and $T$=2s. 
\textbf{Results}: A total of $\mathbb{R}^{5347}$-basis functions are used. The weights of the basis function at all the spatial locations are shown in Fig. \ref{fig:spy_beam}. The notion of identification is the same as in previous examples. It can be seen the weight matrix is highly sparse. The active functions are also portrayed in the figure. The associated parameters are given in Table \ref{tab:ident_param}. The error in the parameters are negligible, and when used for the weighted sum of the Lagrangian at each spatial location, we obtain the actual Lagrangian of the system.

\subsection{Discovery of Hamiltonian from the identified Lagrangian}
Conservation laws are very important in physical and engineering systems. They are associated with underlying symmetries, and therefore they provide fundamental knowledge about physical systems. Examples of conservation laws include the conservation of energy, conservation of momentum, conservation of mass and etc. In this section, we aim to derive the associated conservation laws from the identified Lagrangian. In particular, we compare the associated Hamiltonian from the actual system with the identified Lagrangian system and prove that our proposed approach can successfully discover new conservation laws directly from data. To derive the Hamiltonian from discovered Lagrangian, we used the following equation,
\begin{equation}
    \mathcal{H}(\bm{X}, \dot{\bm{X}}) = \sum^m_{i=1} \frac{\partial \mathcal{L}(X_i, \dot{X}_i)}{\partial \dot{X}_{i}} \dot{X}_{i} - \mathcal{L}(X_i, \dot{X}_i)
\end{equation}

\begin{figure}[ht!]
    \centering
    \includegraphics[width=0.8\textwidth]{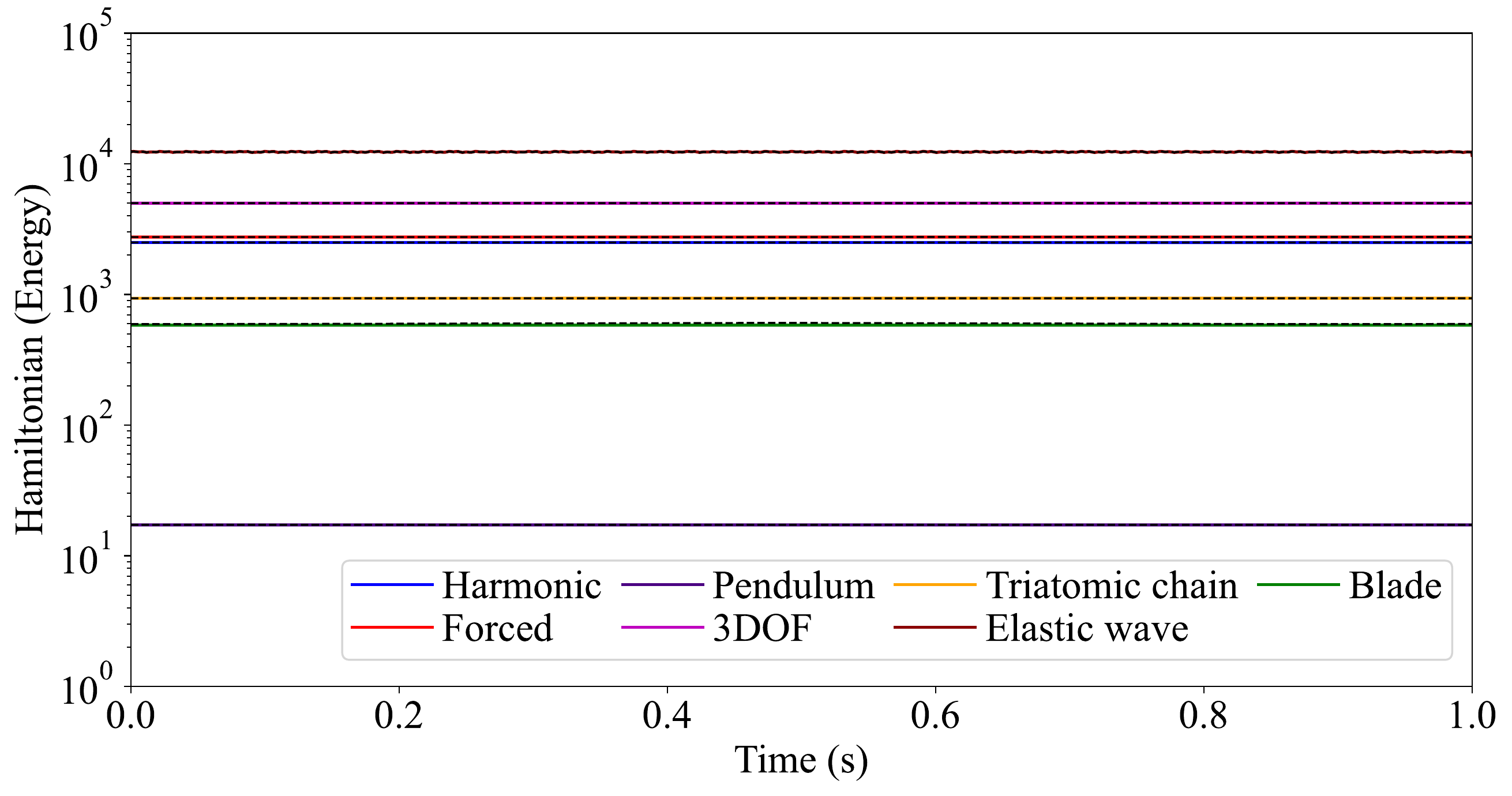}
    \caption{\textbf{Conservation of energy in the example problems}. The solid lines represent the actual energy of the underlying systems, and the dotted lines denote the Hamiltonian of the identified systems. In all the examples, the conservation of energy is discovered. The total energy discovered using the proposed framework matches the Hamiltonian of the actual system.}
    \label{fig:energy}
\end{figure}
The identified Hamiltonian is then compared with the actual Hamiltonian of the system, i.e., the total energy of the system, which is given as $\mathcal{H} = T + V$. The identified Hamiltonians are provided in Table \ref{table:identify}. One can verify that the identified Hamiltonians perfectly match the true ones. For better visualization, we have also portrayed the Hamiltonian, i.e., the conservation of energy in Fig. \ref{fig:energy}. Firstly, we can comment that the energy is conserved in the identified systems. Secondly, we can conclude that the identified energy in the systems is the same as the actual system. This proves our claim that the proposed framework can successfully discover the Lagrangian as well as the conservation laws from only a single observation of systems states.

\subsection{Discovery of equations of motion from identified Lagrangian}
\begin{figure}[ht!]
     \centering
     \begin{subfigure}[b]{0.31\textwidth}
        \centering
        \includegraphics[width=\textwidth]{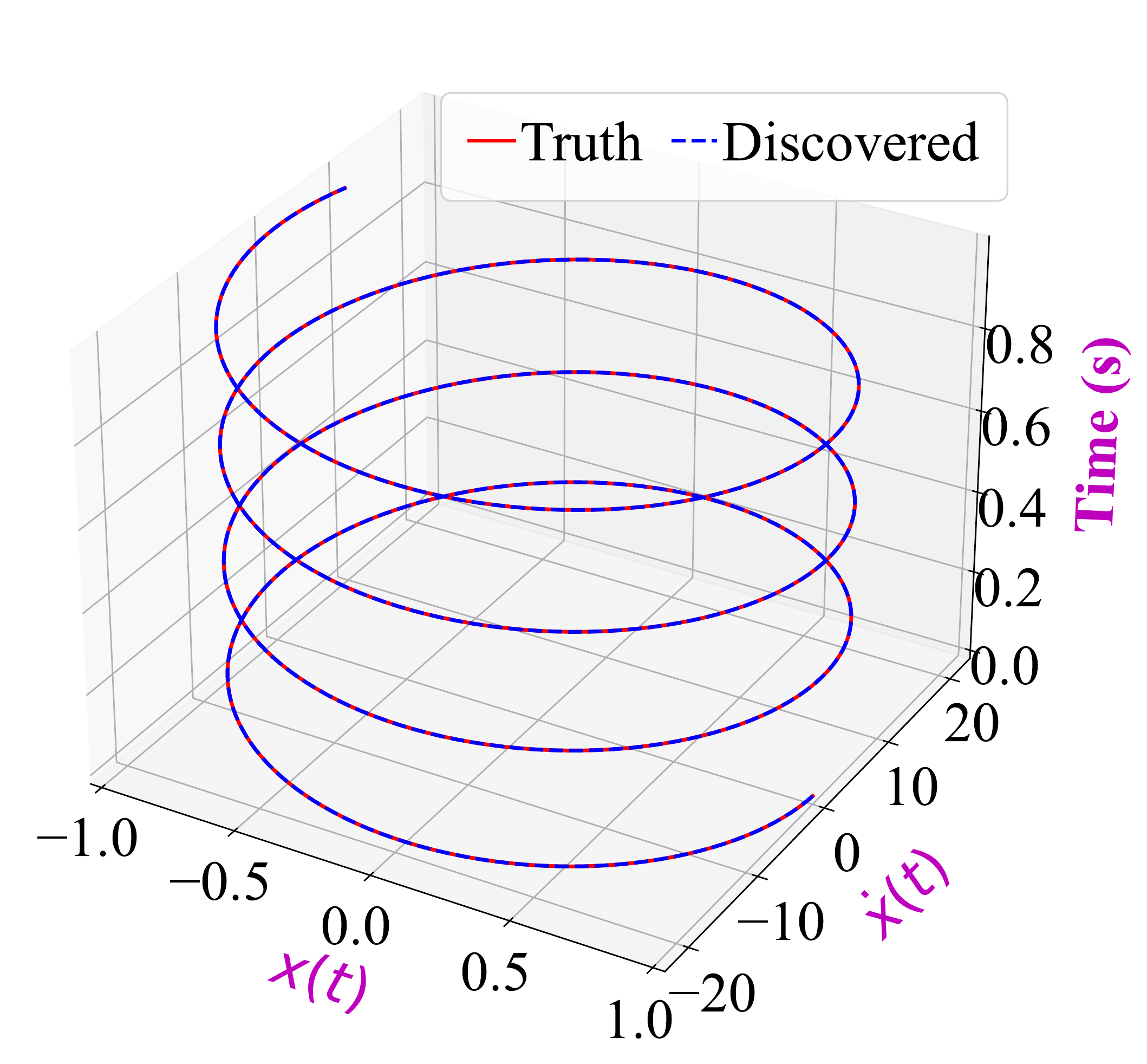}
        \caption{Harmonic oscillator}
        \label{fig:response_linear}
     \end{subfigure}
     % \hfill
     \begin{subfigure}[b]{0.31\textwidth}
        \centering
        \includegraphics[width=\textwidth]{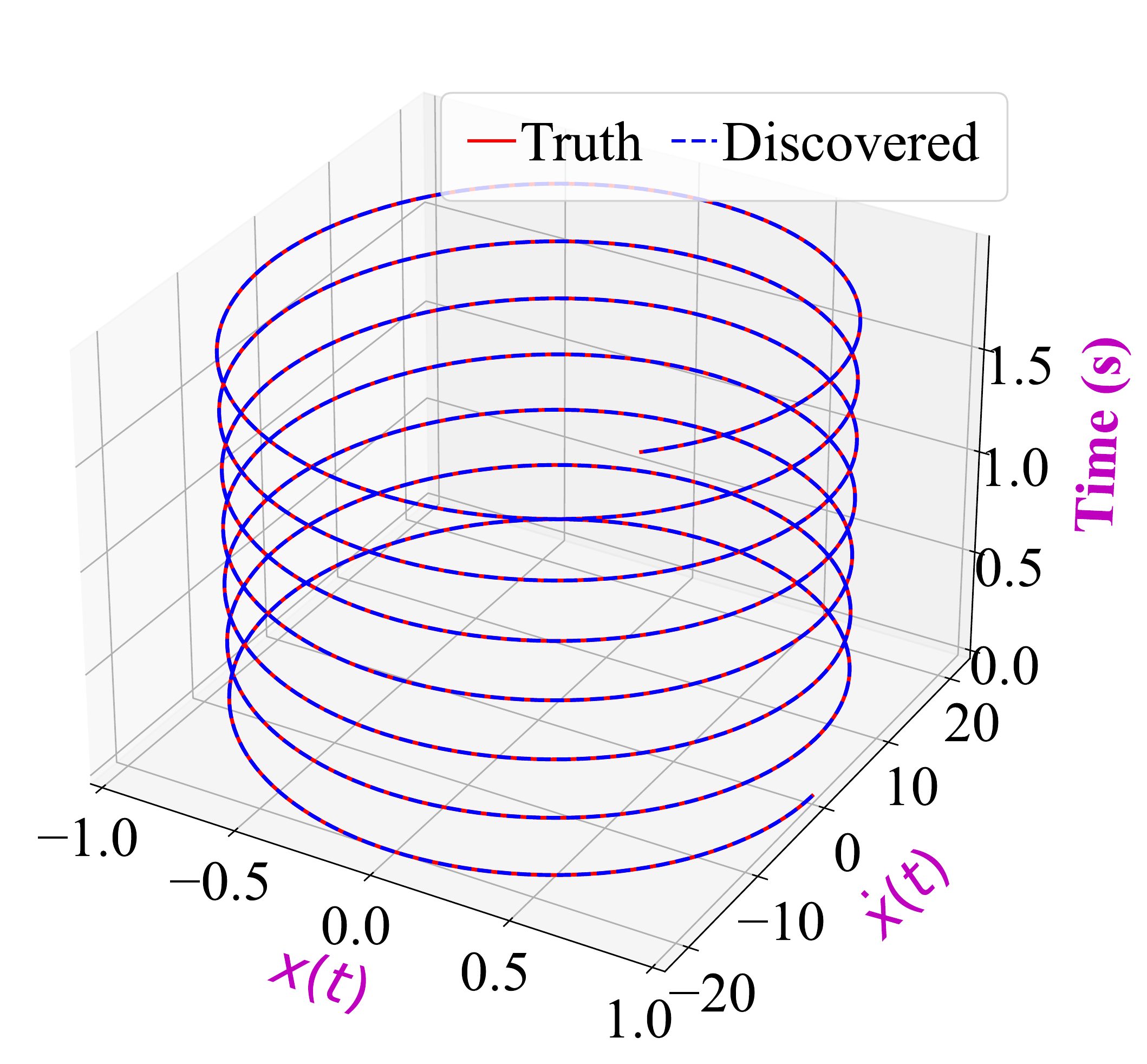}
        \caption{Forced oscillator}
        \label{fig:response_forced}
     \end{subfigure}
     % \hfill
     \begin{subfigure}[b]{0.31\textwidth}
        \centering
        \includegraphics[width=\textwidth]{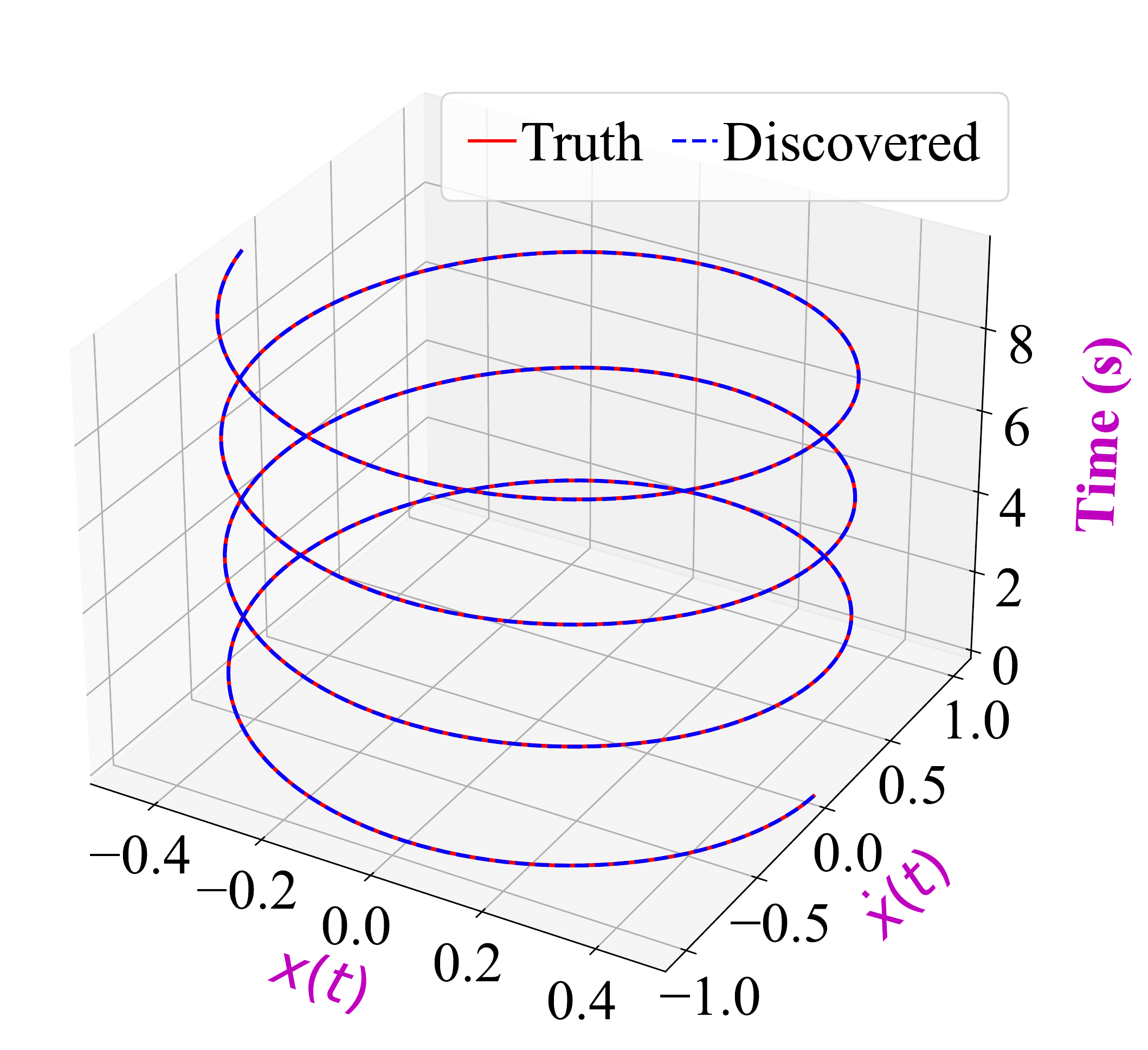}
        \caption{Pendulum}
        \label{fig:response_pendulam}
     \end{subfigure}
     % \hfill
     
     \begin{subfigure}[b]{\textwidth}
        \centering
        \includegraphics[width=0.9\textwidth]{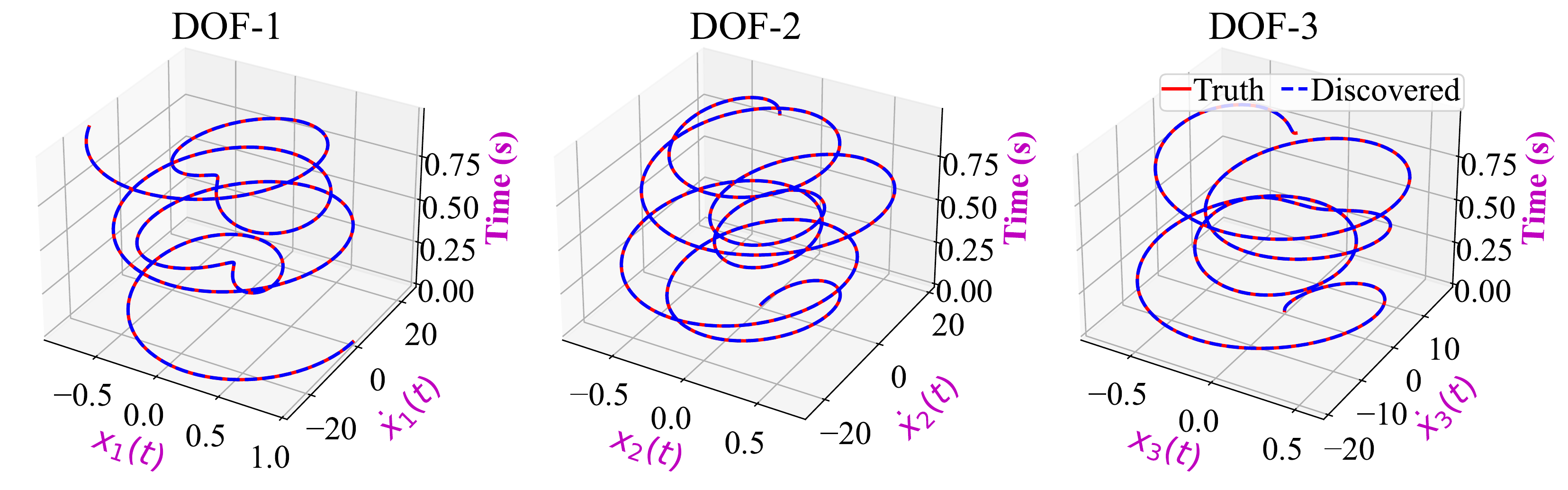}
        \caption{3DOF dynamical system}
        \label{fig:response_3dof}
     \end{subfigure}
     
     \begin{subfigure}[b]{\textwidth}
        \centering
        \includegraphics[width=0.9\textwidth]{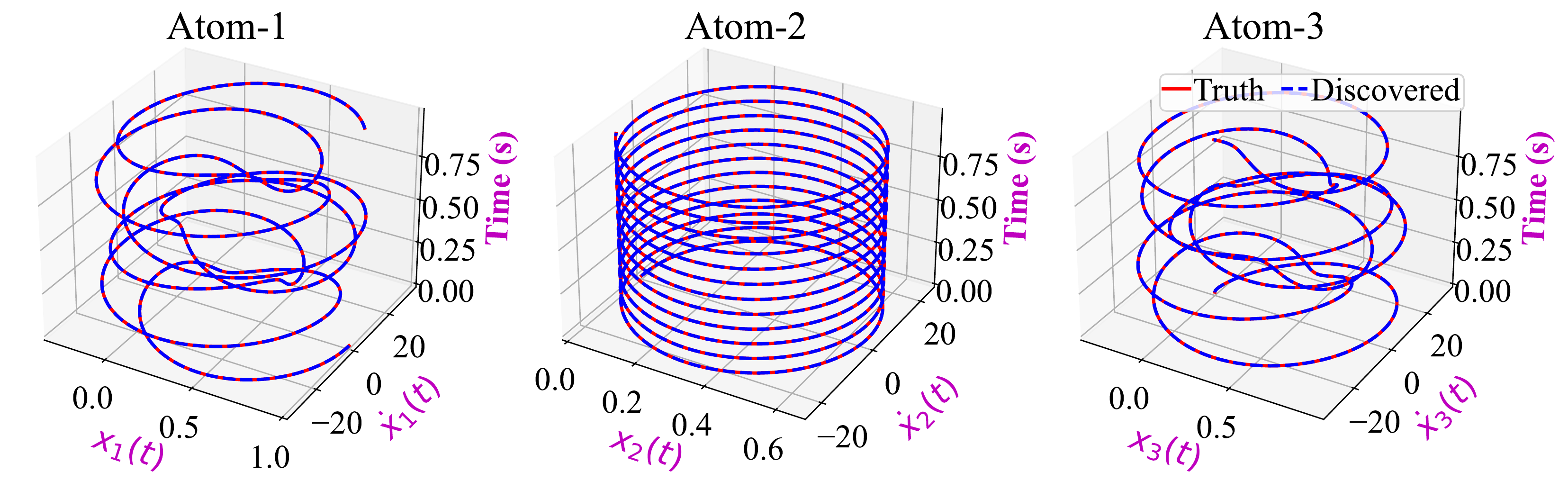}
        \caption{Triatomic molecular chain}
        \label{fig:response_atom}
     \end{subfigure}
        \caption{\textbf{Response of the Lagrangian equation of motion}. The true responses of the actual systems are compared with the responses of the identified Lagrangian equation of motion. In all cases, the solution of identified equation motion completely overlaps the ground truth.}
        \label{fig:response}
\end{figure}

\begin{figure}[ht!]
    \centering
    \includegraphics[width=\textwidth]{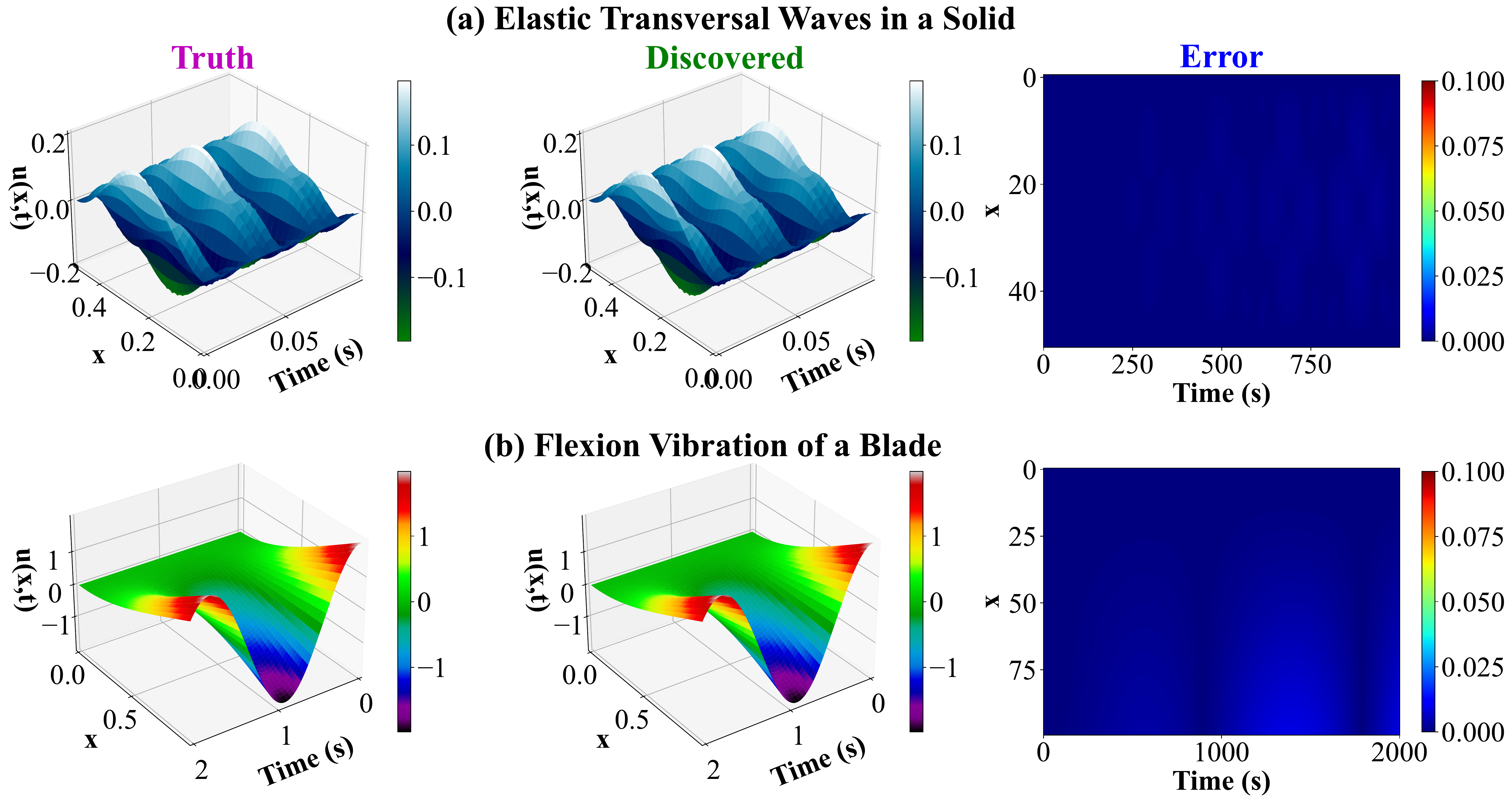}
    \caption{\textbf{Response of the Lagrangian equation of motion of the PDEs}. The solutions of the actual PDEs and the identified Lagrangian PDEs are compared. The error between actual and identified PDEs' solutions is less than 0.25\% in both examples. This indicates that the identified Lagrangians and the derived Lagrangian equations of motion are highly accurate.}
    \label{fig:Response_pde}
\end{figure}

Lagrange’s equations provide a systematic approach to formulate the governing equations of motion of mechanical and structural systems. Instead of force balance, a scalar outlook is taken in the Lagrange approach by expressing the scalar quantities of kinetic and potential energy in terms of generalized coordinates. In this section, we provide enough evidence for the superiority of the proposed framework by constructing the equations of motion from the discovered Lagrangian and then comparing the solutions with their true counterparts. To arrive at the equations of motion, we have applied the  \eqref{eq:lagrange_hom} on the identified Lagrangian. It is to be noted that the exact analytical expressions for the Lagrangian are known (see Table \ref{table:identify}). Thus, when  \ref{eq:lagrange_hom} is applied to the identified Lagrangian, it yields the interpretable equations of motion. This confirms that the proposed framework can not only identify the Lagrangian and conservation laws but is capable of discovering explainable analytical equations of motion from data. The solutions to the derived equations of motion from discovered Lagrangian and actual systems are portrayed in Fig. \ref{fig:response} and \ref{fig:Response_pde}.

In Fig. \ref{fig:response}, the prediction results for ODEs using the identified systems for different initial conditions are portrayed. The solid red lines denote the true solution, and the blue dotted lines represent the solutions of the identified systems. From Fig. \ref{fig:response}(a), \ref{fig:response}(b), and \ref{fig:response}(c) it is evident that the identified systems from the discovered Lagrangian almost exactly overlap the true solutions. Further, from \ref{fig:response}(d) and \ref{fig:response}(e), it can be referred that the identified systems can also approximate the highly complex behavior of the true systems. Similarly, in Fig. \ref{fig:Response_pde}, the predictions on different initial conditions for PDEs are illustrated. The error plots reveal that the discovered systems from the identified Lagrangian can very accurately predict the behavior of actual systems. 

\subsection{Zero shot generalization}
\begin{figure}[ht!]
     \centering
     \begin{subfigure}[b]{0.49\textwidth}
        \centering
        \includegraphics[width=\textwidth]{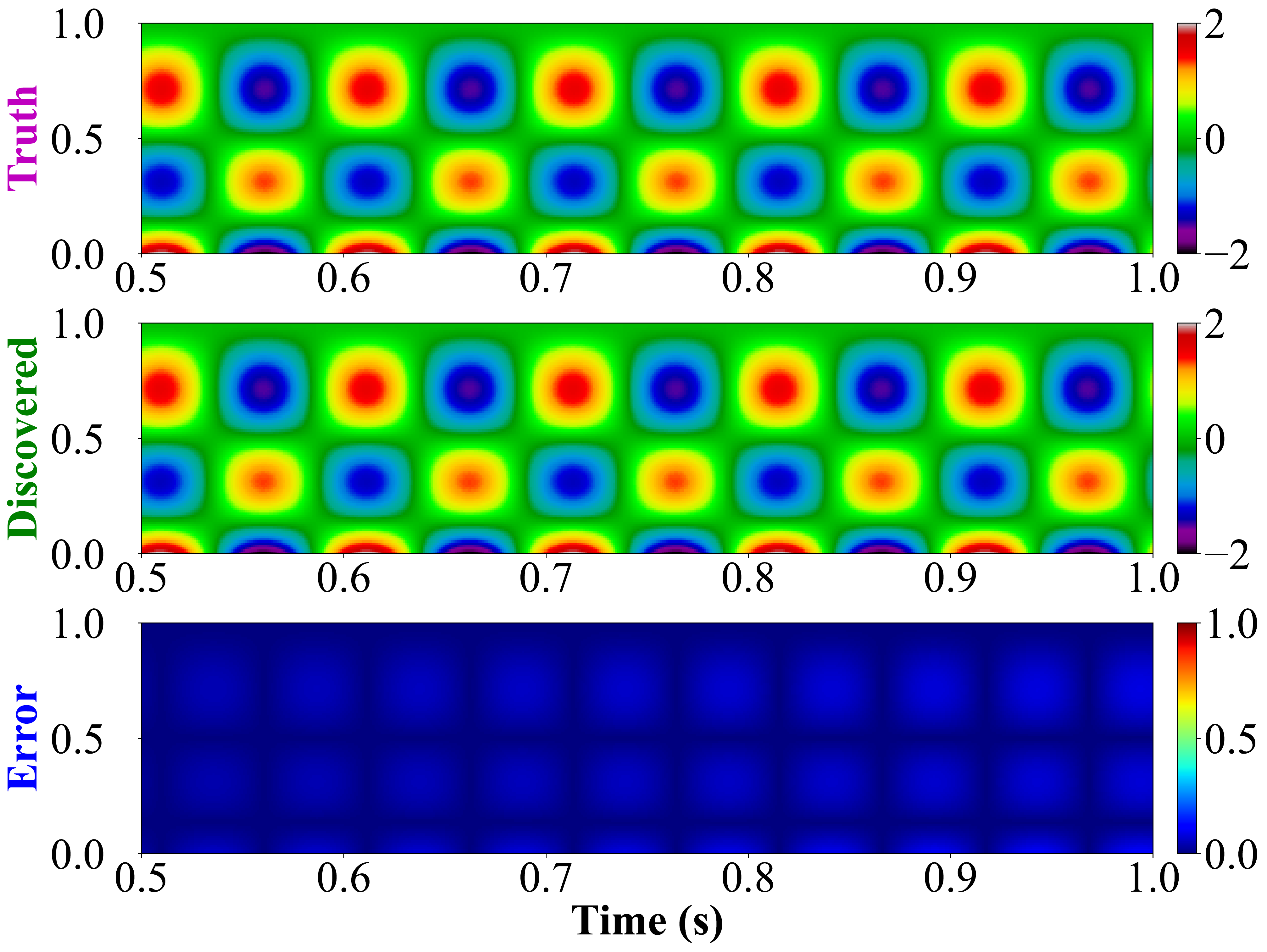}
        \caption{Zero shot generalization}
        \label{fig:zero_shot}
     \end{subfigure}
     \hfill
     \begin{subfigure}[b]{0.49\textwidth}
        \centering
        \includegraphics[width=\textwidth]{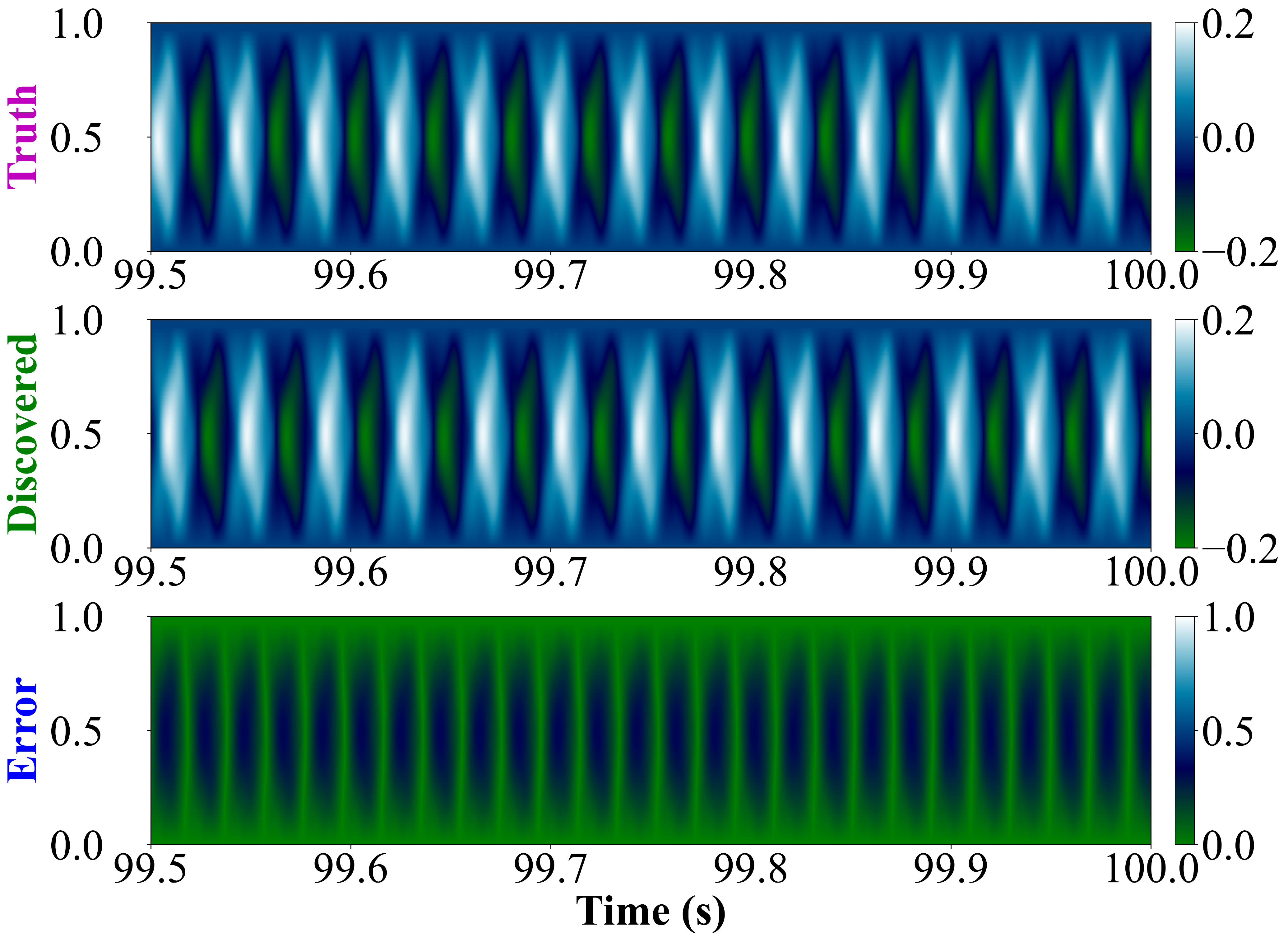}
        \caption{Perpetual predictive capability}
        \label{fig:perpetual}
     \end{subfigure}
        \caption{\textbf{Predictive performance of the proposed framework}. (a) For the discovery of the governing physics of the flexion motion of the blade, the blade is excited using its first natural frequency. In this case, the identified blade physics is solved in its third natural frequency. (b) The governing physics of the transversal motion of the string is identified from 1s of data (recorded at 1000Hz). The identified system is then solved for 100s at the same sampling frequency. In both studies, the responses of the true and identified systems are plotted. In addition, the absolute errors between the solutions are also shown. In both cases, the absolute error is $\ll 1\%$.}
        \label{fig:case1}
\end{figure}

When the governing physics of an underlying physical system is discovered, the purpose is not only limited to the identification of the system. It is often useful to reuse the governing physics for a variety of other purposes, for e.g., testing the stability of the system, prediction on a different condition, numerical modeling, etc. In such cases, the accuracy of the identified system plays an important role. In this section, we consider the motion equation of the traveling wave in the blade (the derived equation from the corresponding discovered Lagrangian) and demonstrate that the motion equation generalizes to other vibrational modes. In particular, we excite the blade using its third natural frequency (note that during the discovery phase, the data are generated by exciting the blade with its first natural frequency) and compare the responses with the actual system. The result is displayed in Fig. \ref{fig:zero_shot}. We see that the error between the responses of the identified and true systems ($\ll 1\%$) is quite negligible.

\subsection{Perpetual predictive capability}
In all the governing physics discovery problems, the prime focus is to capture the actual dynamics of the underlying system rather than to predict the given data. When the actual physics of a physical system is captured, maybe in intrinsic coordinates \cite{tripura2023sparse}, it ensures the underlying system dynamics and allows us to predict for an infinite duration of time with high accuracy. This is particularly important for chaotic systems where small deviations in the governing model cause the system to diverge exponentially. In this section, we consider the identified system of the transversal motion of the string and perform a prediction upto 100s. Note that the Lagrangian was identified using a single observation of system response recorded for 1s (with sampling frequency 10000Hz). The result is compared with the actual response in Fig. \ref{fig:perpetual}. It can be seen that the motion equation derived from the identified sparse Lagrangian not only reproduces the actual system but maintains accuracy below $\ll 1\%$. This clearly indicates the perpetual predictive ability of the proposed framework.

\subsection{Generalization to high-dimensional domains}
As stated in Remark 1, when data are collected from a large number of sensor locations (e.g., spatial discretization of PDEs, where the dimension of the discretized system may become in the order of tens of thousands to billions of variables), the state dimension $m$ becomes prohibitively large. In these cases, the physics discovery of underlying systems may become computationally heavy due to the exponential growth of the dictionary dimension. However, the proposed framework provides a remedy in which we can identify the underlying Lagrangian from a subset of the measurements and then can analytically extend to $m$-dimensions. In this section, we demonstrate the ability of the proposed framework to generalize the Lagrangian identified from a relatively small domain to the Lagrangian density of the whole domain. In particular, we consider the Lagrangian of the triatomic molecule and aim to generalize it to an $m$-link atomic chain (see Table \ref{tab:ident_param} and \ref{table:identify}).
\begin{figure}[!ht]
    \centering
    \includegraphics[width=\textwidth]{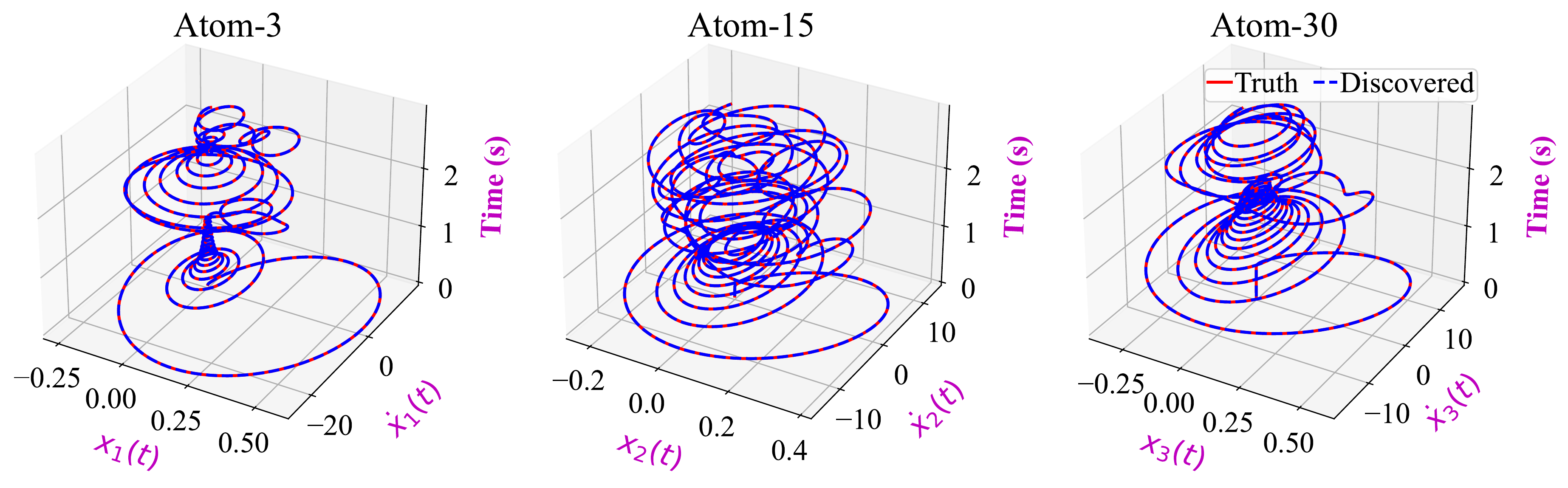}
    \caption{\textbf{High-dimensional generalization ability of the proposed framework}. An atomic chain with 30 molecules is generated from the Lagrangian of the triatomic molecule. The responses for a few atoms of the identified chain are plotted. The generalized system accurately identifies the correct vibrational modes of the atomic chain.}
    \label{fig:case_general}
\end{figure}

In the $n$-link chain, we consider an elementary chain consisting of light atoms with mass $m$ and heavy atoms with mass $\tilde{m}$. Let the associated displacements be $X_j$ and $\tilde{X}_j$, respectively. Then the kinetic energy of the elementary chain is $T_j=\frac{1}{2} (m\dot{X}_j^2+ \tilde{m}\dot{X}_j^2 )$.
From the triatomic chain, we can find that the net elongation between the $j^{th}$ light and heavy atom is $(\tilde{X}_j - X_j)$. The corresponding potential energy is $\frac{1}{2} k(\tilde{X}_j-X_j)^2$. Also, between the heavy atom $j$ and the light atom $j+1$, the net elongation is $(X_{j+1}-\tilde{X}_j)$ to which corresponds a potential energy $\frac{1}{2} k(X_{j+1}-\tilde{X}_j)^2$. The Lagrangian of the chain is then found by computing the difference between the total kinetic energy and total potential energy as follows,
\begin{equation}
    \mathcal{L}(\bm{X}, \dot{\bm{X}}) = \sum_j^n{\left[\frac{1}{2} m\dot{X}_j^2 + \frac{1}{2} \tilde{m}\dot{\tilde{X}}_j^2 - \frac{1}{2} k\left(\tilde{X}_j-X_j\right)^2-\frac{1}{2} k\left(X_{j+1}-\tilde{X}_j\right)^2\right]} .
\end{equation}
For this study, we use a chain of $m=30$ and keep the same parameters of the system. The results are summarized in Fig. \ref{fig:case_general}. It is evident that the responses of the generalized chain and the actual system are indistinguishable. The generalized system reproduced the dynamics of the actual system very well.

\subsection{Noise sensitivity}
In this section, we test the performance of the proposed framework against the measurement noise. In this investigation, we corrupt the single observation of data with zero mean Gaussian noise. For brevity, we limit this study to the (i) Harmonic oscillator (Ex. \ref{eq:free}), (ii) 3DOF dynamical system (Ex. \ref{example_4}), and Triatomic molecule (Ex. \ref{example_5}). A total of 5 levels of noise are considered, where each level indicates a certain percentage of the standard deviation of the actual data. For the sake of comparison, we first check whether the correct dictionary basses are identified. In the presence of any other functions (even if the correct basses are identified), we discard the identified models since, in the presence of false basses, the generalization ability of discovered model decreases. If the identified sparse model exactly matches with the actual system, we compute an error $e_{\mathcal{L}} = \|\mathcal{L}_a - \mathcal{L}^* \|_2 / \|\mathcal{L}_a\|_2$ to test the accuracy of the model. Here $\mathcal{L}_a$ and $\mathcal{L}^*$ represent the actual and identified Lagrangian, respectively. The corresponding results are illustrated in Table \ref{tab:noise}. We observe that the proposed framework failed to identify the exact analytical model of the Lagrangian of the triatomic molecule and 3DOF system when noise $\zeta \ge 4$ and $\zeta \ge 5$, respectively. From the study, we conclude that the proposed framework works successfully when the data are corrupted with noise levels between $3-4\%$. 

\begin{table}[!ht]
    \centering
    \caption{Noise sensitivity of the proposed framework}
    \begin{tabular}{llllllll}
        \toprule
         \multirow{2}{*}{ Noise level-$\zeta$ (\%)}& \multicolumn{3}{c}{Discovery of correct basis} && \multicolumn{3}{c}{Relative Error, $e_{\mathcal{L}}$ (\%)} \\ \cline{2-4} \cline{6-8}
         & Harmonic OSC. & Triatomic mol. & 3DOF && Harmonic OSC. & Triatomic mol. & 3DOF \\
         \midrule
         No noise & \textcolor{blue}{Yes} & \textcolor{blue}{Yes} & \textcolor{blue}{Yes} && 0.0280 & 0.3916 & 0.1554 \\
         \hdashline
         1 & \textcolor{blue}{Yes} & \textcolor{blue}{Yes} & \textcolor{blue}{Yes} && 0.0096 & 0.4756 & 0.2139 \\
         2 & \textcolor{blue}{Yes} & \textcolor{blue}{Yes} & \textcolor{blue}{Yes} && 0.0245 & 0.5574 & 0.9455 \\
         3 & \textcolor{blue}{Yes} & \textcolor{blue}{Yes} & \textcolor{blue}{Yes} && 0.1956 & 1.1649 & 1.6566 \\
         4 & \textcolor{blue}{Yes} & \textcolor{red}{No} & \textcolor{blue}{Yes} && 0.5076 & -- & 3.1050 \\
         5 & \textcolor{blue}{Yes} & \textcolor{red}{No} & \textcolor{red}{No} && 1.1394 & -- & -- \\
         \bottomrule
    \end{tabular}
    \label{tab:noise}
\end{table}

\section{Discussions}\label{sec:conclusion}
We have proposed a novel framework for the discovery of Lagrangian of systems from a single observation of data using sparse least-squares regression. In particular, we combine the Lagrangian formalism with sequential threshold least-squares to discover the Lagrangian. Briefly, we collected several candidate functions for the Lagrangian inside a dictionary and then utilized the sparse regression to identify a combination of candidate functions that satisfy the principle of minimal action. We have shown that the resulting framework is highly robust and can learn exact sparse forms of the Lagrangian of systems (both rigid and flexible systems) from data alone, whereas the available Lagrangian discovery schemes are black-box in nature. We have also shown that the proposed framework can recover the governing equations of motion (for systems described by ODEs and PDEs) without extra effort. Through the Legendre transformation, it is also possible to derive conservation laws from the proposed framework. From the discovery of Lagrangian to the identification of hidden conservation laws to distilling governing equations of motion, the proposed framework utilizes only a single set of observations of data. The robustness of the proposed framework also lies in its ability to generalize to different initial conditions and the ability to predict the actual system dynamics for an infinite duration of time. When the underlying observation states are humongous, the proposed framework also provides a way to learn the Lagrangian from a relatively small subset of the observations. The fidelity is demonstrated on a sufficient number of systems of ODEs and PDEs, including various case studies. The results are highly encouraging.

To our knowledge, there is only one major limitation of the proposed work. We note that in our proposed framework, the accuracy of the discovered Lagrangian depends on the right choice of candidate function in the dictionary. However, in the presence of dissipative forces \cite{zhong2020dissipative}, the candidate function may be entangled between states in a very complicated fashion. In such cases, choosing the right basis function may not be possible, and the discovered Lagrangian may generalize beyond some time frame. However, having prior knowledge may alleviate the problem to some degree.

Considering the above limitation, we believe the proposed framework can automate, accelerate, and guide the discovery of new physics and associated conservation the way a human would discover, especially where the human cognitive capabilities are met limits. The discovery of the exact analytical form of Lagrangian rather than a direct discovery of governing equations of motion will also enable the knowledge-based design of future technologies. An immediate application would include the integration of the proposed approach for learning and control of dynamical systems \cite{gupta2019general,lutter2019deep}.

\appendix

\section*{Acknowledgements} 
T. Tripura acknowledges the financial support received from the Ministry of Education (MoE), India, in the form of the Prime Minister’s Research Fellowship (PMRF). S. Chakraborty acknowledges the financial support received from Science and Engineering Research Board (SERB) via grant no. SRG/2021/000467 and seed grant received from IIT Delhi.

\section*{Declarations}

% \subsection*{Funding} The corresponding author received funding from IIT Delhi in form of seed grant.

\subsection*{Conflicts of interest} The authors declare that they have no conflict of interest.

\subsection*{Availability of data and material} The datasets generated during and/or analyzed during the current study are available from the corresponding author upon reasonable request.

\subsection*{Code availability} The Python codes written for this work are available from the corresponding author upon reasonable request.

% % Bibliography
% \bibliographystyle{unsrt}  
% \bibliography{mybibfile}  

\end{document}